\documentclass[runningheads]{llncs}
\usepackage{graphicx}
\usepackage{soul}
\usepackage{color}
\soulregister\cite7
\usepackage{amsmath,amssymb} 

\usepackage[width=122mm,left=12mm,paperwidth=146mm,height=193mm,top=12mm,paperheight=217mm]{geometry}

\begin{document}
\pagestyle{headings}
\mainmatter

\title{Built-in Foreground/Background Prior for Weakly-Supervised Semantic Segmentation} 

\titlerunning{Built-in Fg/Bg Prior for Weakly-Supervised Semantic Segmentation}

\authorrunning{Saleh, Ali Akbarian, Salzmann, Petersson, Gould, and Alvarez}

\author{Fatemehsadat Saleh$^{1,2}$, Mohammad Sadegh Ali Akbarian$^{1,2}$, Mathieu Salzmann$^{1,3}$, Lars Petersson$^{1,2}$, Stephen Gould$^{1}$, and Jose M. Alvarez$^{1,2}$}


\institute{$^{1}$The Australian National University (ANU),\\
\email{stephen.gould@anu.edu.au}\\
$^{2}$Commonwealth Science and Industrial Research Organization (CSIRO), \\
\email{\{fatemehsadat.saleh,mohammadsadegh.aliakbarian,  lars.petersson, jose.alvarezlopez\}@data61.csiro.au}\\
$^{3}$CVLab, EPFL, Switzerland\\
\email{mathieu.salzmann@epfl.ch}\\
}

\maketitle

\begin{abstract}
Pixel-level annotations are expensive and time consuming to obtain. Hence,  weak supervision using only image tags could have a significant impact in semantic segmentation. Recently, CNN-based methods have proposed to fine-tune pre-trained networks using image tags. Without additional information, this leads to poor localization accuracy. This problem, however, was alleviated by making use of objectness priors to generate foreground/background masks. Unfortunately these priors either require training pixel-level annotations/bounding boxes, or still yield inaccurate object boundaries. Here, we propose a novel method to extract markedly more accurate masks from the pre-trained network itself, forgoing external objectness modules. This is accomplished using the activations of the higher-level convolutional layers, smoothed by a dense CRF. We demonstrate that our method, based on these masks and a weakly-supervised loss, outperforms the state-of-the-art tag-based weakly-supervised semantic segmentation techniques. Furthermore, we introduce a new form of inexpensive weak supervision yielding an additional accuracy boost.

\end{abstract}

\section{Introduction}

Semantic scene segmentation, i.e., assigning a class label to every pixel in an input image, has received growing attention in the computer vision community, with accuracy greatly increasing over the years~\cite{FCN,deconvnet,DeepLab,crf-rnn,zoomout,deephierarchy}. In particular, fully-supervised approaches based on Convolutional Neural Networks (CNNs) have recently achieved impressive results~\cite{farabetSeg,FCN,DeepLab,deconvnet,crf-rnn}. Unfortunately, these methods require large amounts of training images with pixel-level annotations, which are expensive and time-consuming to obtain.
Weakly-supervised techniques have therefore emerged as a solution to address this limitation~\cite{graphbasedseg,tellme,multiimagemodel,segmentvarious,whatsthepoint,weaklyandsemi,MIL,cliqueseg}. These techniques rely on a weaker form of training annotations, such as, from weaker to stronger levels of supervision, image tags~\cite{MIL,whatsthepoint,fromimgtopixel,CCNN}, information about object sizes~\cite{CCNN}, labeled points or squiggles~\cite{whatsthepoint} and labeled bounding boxes~\cite{weaklyandsemi,boxsup}. In the current Deep Learning era, existing weakly-supervised methods typically start from a network pre-trained on an object recognition dataset (e.g., ImageNet~\cite{ILSVRC}) and fine-tune it using segmentation losses defined according to the weak annotations at hand~\cite{fromimgtopixel,CCNN,weaklyandsemi,whatsthepoint,MIL}.

In this paper, we are particularly interested in exploiting one of the weakest levels of supervision, i.e., image tags, which is a rather inexpensive attribute to annotate and thus more common in practice (e.g., Flickr~\cite{flickr}). Image tags simply determine which classes are present in the image without specifying any other information, such as the location of the objects. In this extreme setting, a naive weakly-supervised segmentation algorithm will typically yield poor localization accuracy. Therefore, recent works~\cite{fromimgtopixel,whatsthepoint,learnTosegwithimagelevelanno} have proposed to make use of objectness priors~\cite{objectness,BING,MCG,CPMC}, which provide each pixel with a probability of being an object. In particular, these methods have exploited existing objectness algorithms, such as~\cite{objectness,BING,MCG}, with the drawback of introducing external sources of potential error. Furthermore,~\cite{objectness} typically only yields a rough foreground/background estimate, and~\cite{BING,MCG} rely on additional training data with pixel-level annotations.

Here, by contrast, we introduce a Deep Learning approach to weakly-supervised semantic segmentation where the localization information is directly extracted from the network itself. Our approach relies on the following intuition:
One can expect that a network trained for the task of object recognition extracts features that focus on the objects themselves, and thus has hidden layers with units firing up on foreground objects, but not on background regions. A similar intuition was also recently explored for other tasks, such as object localization~\cite{ObjectLocalForFree} and  detection~\cite{ObjectEmerge}.
Starting from a fully-convolutional network pre-trained on ImageNet, we therefore propose to extract a foreground/background mask by directly exploiting the unit activations of some of the hidden layers in the network.

\begin{figure}[t!]
\centering
\includegraphics[width=0.9\textwidth]{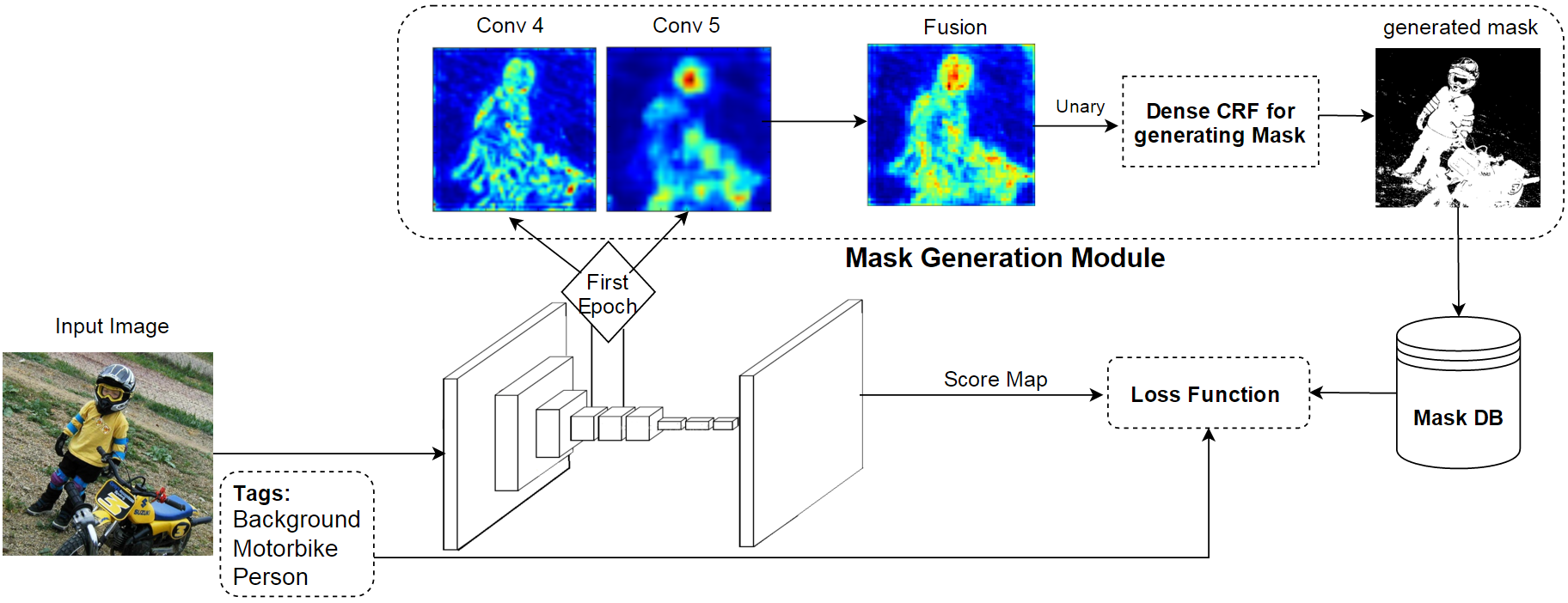}
\caption{Our weakly-supervised network with built-in foreground/background prior.}
\label{fig:intro}
\end{figure}

In particular, we focus on the fourth and fifth convolution layers of the VGG-16 pre-trained network~\cite{VGG16}, which provide higher-level information than the first three layers, such as highlighting complete objects or object parts. We then make use of a fully-connected Conditional Random Field (CRF) to smooth out this information and generate a foreground/background mask. We finally incorporate the resulting masks in our network via a weakly-supervised loss. The resulting masks can also be thought of as a form of objectness measure. While several CNN-based approaches have proposed to learn objectness, or saliency measures from annotations~\cite{deepproposal,deepbox,HARF}, to the best of our knowledge, our approach is the first extract this information directly from the hidden layer activations of a segmentation network, and employ the resulting masks as localization cues for weakly-supervised semantic segmentation.
Ultimately, our model, illustrated by Fig.~\ref{fig:intro}, can therefore be thought of as a weakly-supervised segmentation network with built-in foreground/background prior.

We demonstrate the benefits of our approach on two datasets (Pascal VOC 2012~\cite{pascal} and a subset of Flickr (MIRFLICKR-1M)~\cite{flickr}).
Our experiments show that our approach outperforms the state-of-the-art methods that use image tags only, and even some methods that leverage additional supervision, such as object size information~\cite{CCNN} and point supervision~\cite{whatsthepoint}. Furthermore, we extend our framework to incorporate some additional, yet cheap, supervision, taking the form of asking the user to select the best foreground/background mask among several automatically generated candidates. Our experiments reveal that this additional supervision only costs the user roughly 2-3 seconds per image and yields another significant accuracy boost over our tags-only results.

\section{Related work}
Weakly-supervised semantic segmentation has attracted a lot of attention, because it alleviates the painstaking process of manually generating pixel-level training annotations. Over the years, great progress~\cite{tellme,multiimagemodel,segmentvarious,whatsthepoint,weaklyandsemi,MIL,fromimgtopixel,CCNN,boxsup,structuredlearning} has been made. In particular, recently, Convolutional Neural Networks have been applied to the task of weakly-supervised segmentation with great success. In this section, we discuss these CNN-based approaches, which are the ones most related to our work.

The work of~\cite{MIL} constitutes the first method to consider fine-tuning a pre-trained CNN using image-level tags only within a weakly-supervised segmentation context.
This approach relies on a simple Multiple Instance Learning (MIL) loss to account for image tags during training. While this loss improves segmentation accuracy over a naive baseline, this accuracy remains relatively low, due to the fact that no other prior than image tags is employed. By contrast,~\cite{weaklyandsemi} incorporates an additional prior in the MIL framework in the form of an adaptive foreground/background bias. This bias significantly increases accuracy, which~\cite{weaklyandsemi} shows can be further improved by introducing stronger supervision, such as labeled bounding boxes. Importantly, however, this bias is data-dependent and not trivial to re-compute for a new dataset. Furthermore, the results remain inaccurate in terms of object localization. In~\cite{CCNN}, weakly-supervised segmentation is formulated as a constrained optimization problem, and an additional prior modeling the size of objects is introduced. This prior relies on thresholds determining the percentage of the image area that certain classes of objects can occupy, which again is problem-dependent. More importantly, and as in~\cite{weaklyandsemi}, the resulting method does not exploit any information about the location of objects, and thus yields poor localization accuracy.

To overcome this weakness, some approaches~\cite{fromimgtopixel,whatsthepoint,learnTosegwithimagelevelanno} have proposed to exploit the notion of objectness. In particular,~\cite{fromimgtopixel} makes use of a post-processing step that smoothes their initial segmentation results using the object proposals obtained by BING~\cite{BING} or MCG~\cite{MCG}. While it improves localization, being a post-processing step, this procedure is unable to recover from some mistakes made by the initial segmentation. By contrast,~\cite{whatsthepoint,learnTosegwithimagelevelanno} directly incorporate an objectness score~\cite{objectness,MCG} in their loss function. While accounting for objectness when training the network indeed improves segmentation accuracy, the whole framework depends on the success of the external objectness module, which, in practice, only produces a coarse heat map and does not accurately determine the location and shape of the objects (as evidenced by our results in supplementary materials).

Note that BING and MCG have been trained from PASCAL \textit{train} images with full pixel-level annotations or bounding boxes, and thus~\cite{fromimgtopixel,learnTosegwithimagelevelanno} inherently makes use of stronger supervision than our approach. Here, instead of relying on an external objectness method, we leverage the intuition that, within its hidden layers, a network pre-trained for object recognition should already have learned to focus on the object themselves. This lets us develop a foreground/background mask directly from the information built into the network, which we empirically show provides a more accurate object localization prior.
A relevant idea is also presented in an arxiv paper~\cite{STC} which is further evidencing the popularity and importance of this research trend.

\section{Our Approach}
In this section, we introduce our approach to weakly-supervised semantic segmentation. After briefly discussing the CNN architecture that we use, we present our approach to extracting a foreground/background mask directly from the network itself. We then introduce our weakly-supervised learning algorithm that leverages this foreground/background information, and finally discuss our novel way to introduce additional weak supervision in the process.

\subsection{Network Architecture}
\label{sec:architecture}
\begin{figure}[t!]
\centering
\includegraphics[width=0.9\textwidth]{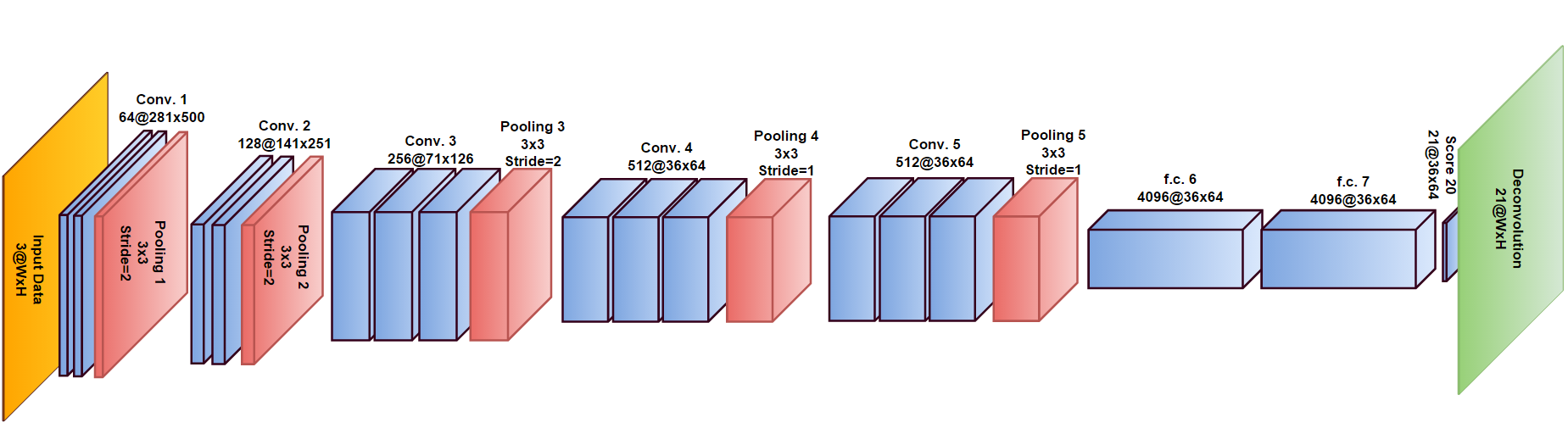}
\caption{Network Architecture: Fully convolutional neural network, derived from the VGG-16 network. We employ a receptive field of 128 pixels and a stride of 8. }
\label{fig:net-arch}
\end{figure}

As most recent weakly-supervised semantic segmentation algorithms~\cite{whatsthepoint,weaklyandsemi,MIL,fromimgtopixel,CCNN}, and as shown in Fig.~\ref{fig:net-arch}, our architecture is based on the VGG-16-layer network~\cite{VGG16}, whose weights were trained on ImageNet for the task of object recognition. Following the fully-convolutional approach~\cite{FCN}, all fully-connected layers are converted to convolutional layers, and the final classifier replaced with a \(1\times 1\) convolution layer with $N$ channels, where $N$ represents the number of classes of the problem. As a modification to this fully convolutional network which has a stride of 32, inspired from~\cite{DeepLab}, we use a stride of 8 and also a smaller receptive field (128 pixels), which has proven to be effective in practice in weakly-supervised semantic segmentation~\cite{weaklyandsemi}.
At the end of the network, we add a deconvolution layer to up-sample the output of the network to the size of the input image. In short, the network takes an image of size \(W\times H\) as input and generates an \(N\times W\times H\) output encoding a score for each pixel and for each class.

\subsection{Built-in Foreground/Background Model}
\label{sec:fb}
We now introduce our approach to extracting a foreground/background mask directly from our network. In Section~\ref{sec:learning}, we show how this mask can be employed for weakly-supervised semantic segmentation.

\begin{figure}[t!]
\centering
\tiny
\begin{tabular}{c c c c c c c c c  }
Image & \(1^{st}\) Conv. & \(2^{nd}\) Conv. & \(3^{rd}\) Conv. & \(4^{th}\) Conv. & \(5^{th}\) Conv. & Fusion & Our mask & G.T\\
\includegraphics[width=.09\textwidth]{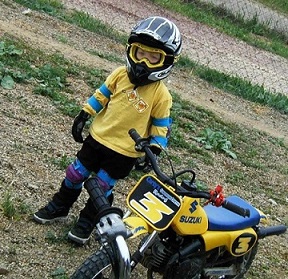} &
\includegraphics[width=.09\textwidth]{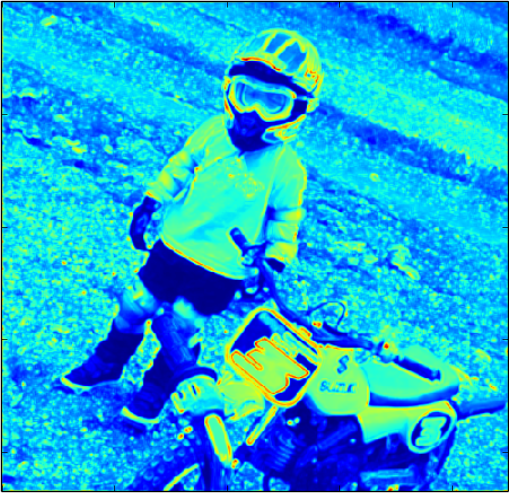} &
\includegraphics[width=.09\textwidth]{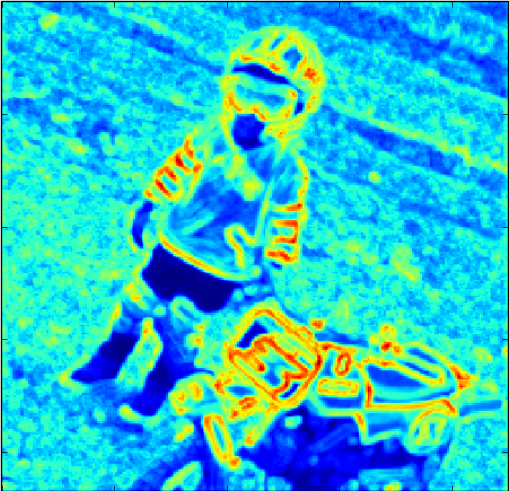} &
\includegraphics[width=.09\textwidth]{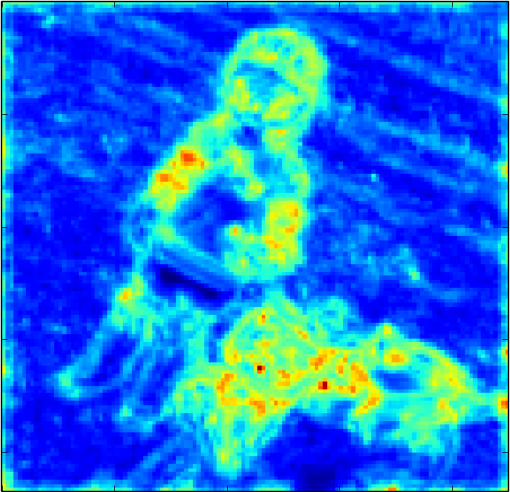} &
\includegraphics[width=.09\textwidth]{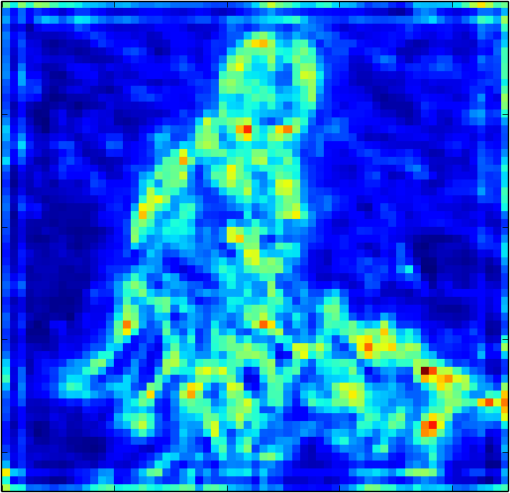} &
\includegraphics[width=.09\textwidth]{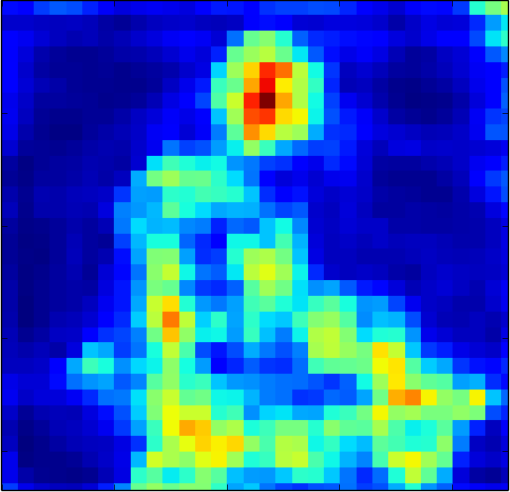} &
\includegraphics[width=.09\textwidth]{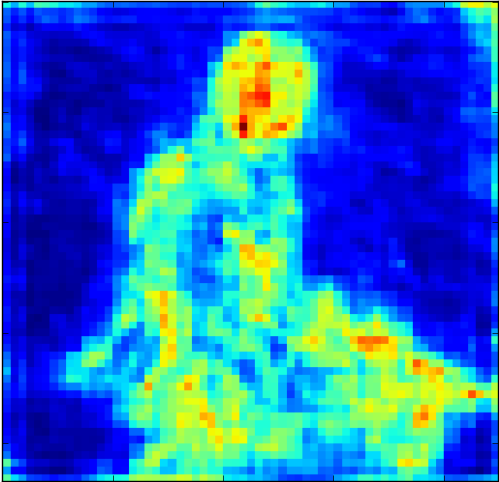} &
\includegraphics[width=.09\textwidth]{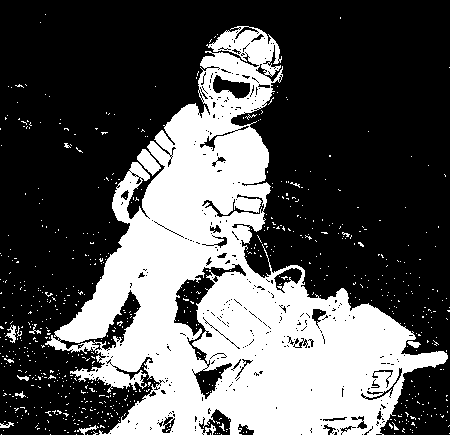} &
\includegraphics[width=.09\textwidth]{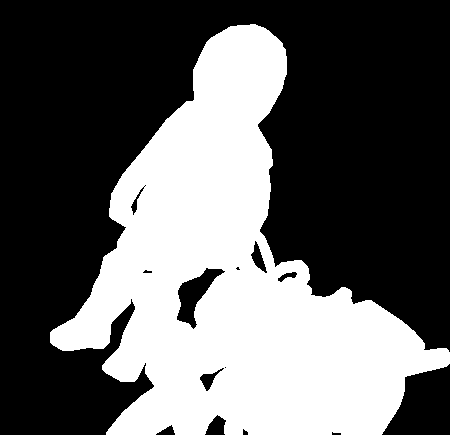}\\

\includegraphics[width=.09\textwidth]{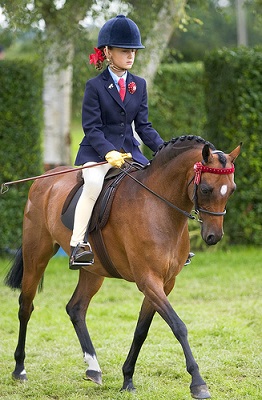} &
\includegraphics[width=.09\textwidth]{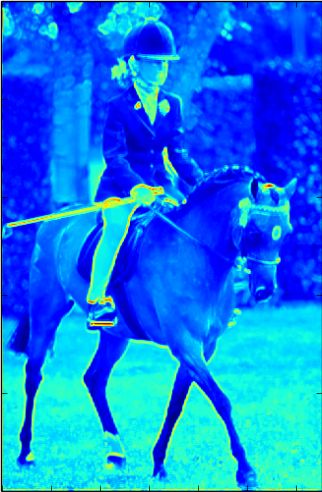} &
\includegraphics[width=.09\textwidth]{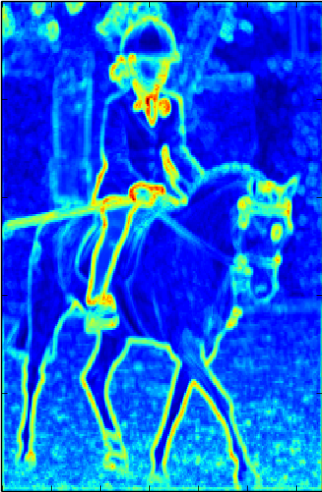} &
\includegraphics[width=.09\textwidth]{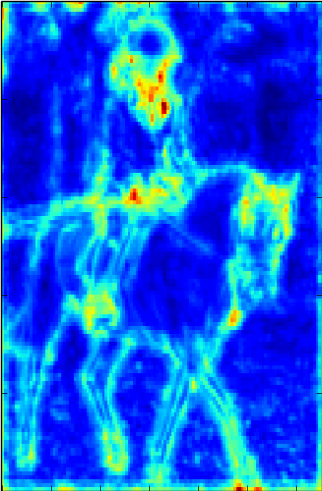} &
\includegraphics[width=.09\textwidth]{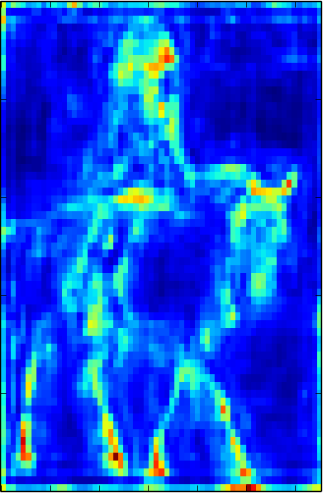} &
\includegraphics[width=.09\textwidth]{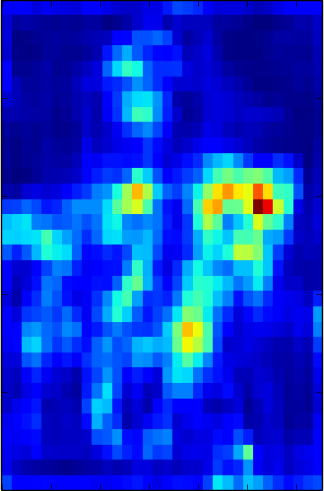} &
\includegraphics[width=.09\textwidth]{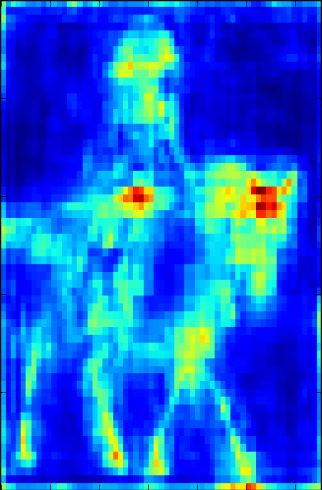} &
\includegraphics[width=.09\textwidth]{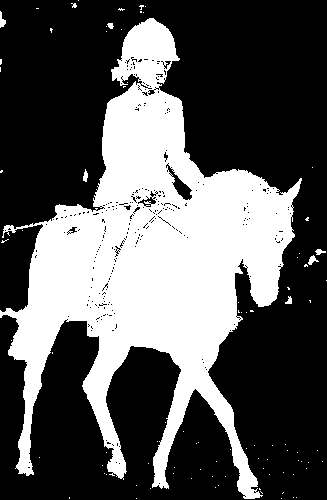} &
\includegraphics[width=.09\textwidth]{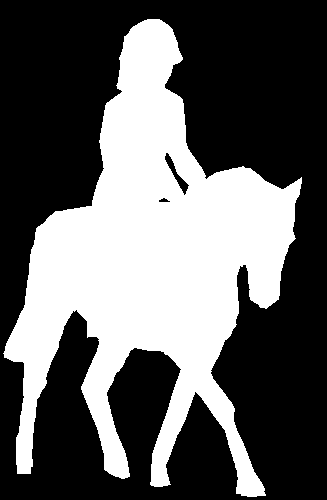} \\

\includegraphics[width=.09\textwidth]{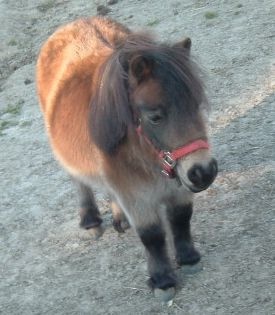} &
\includegraphics[width=.09\textwidth]{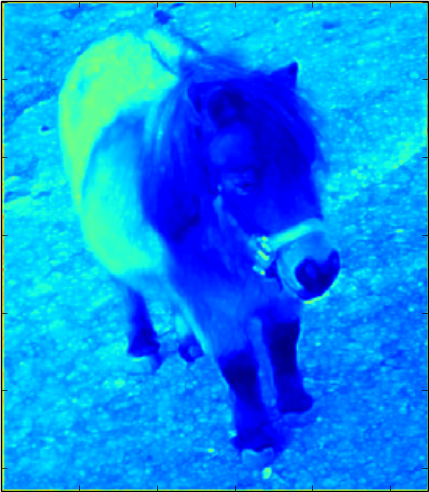} &
\includegraphics[width=.09\textwidth]{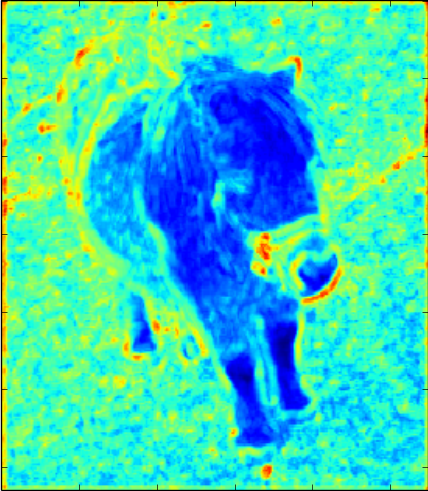} &
\includegraphics[width=.09\textwidth]{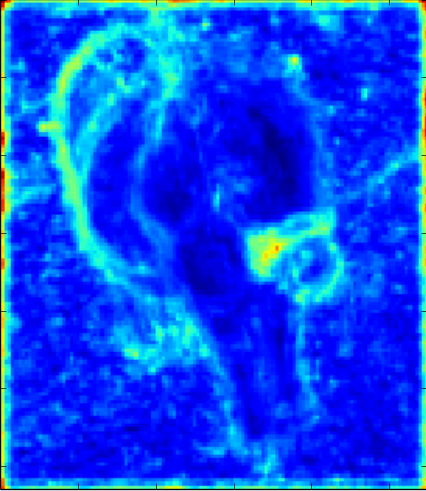} &
\includegraphics[width=.09\textwidth]{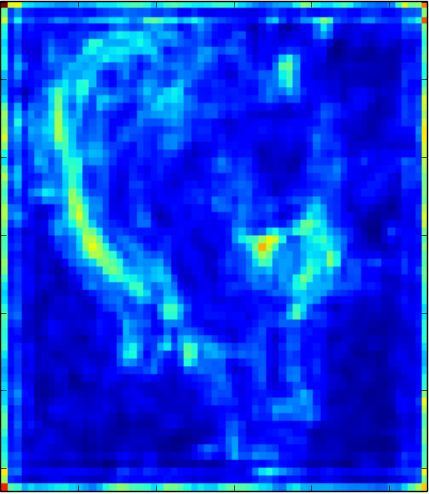} &
\includegraphics[width=.09\textwidth]{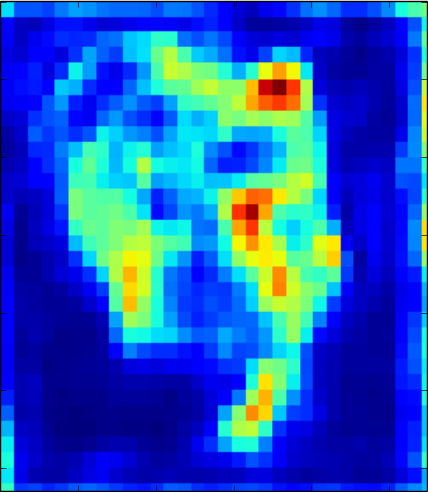} &
\includegraphics[width=.09\textwidth]{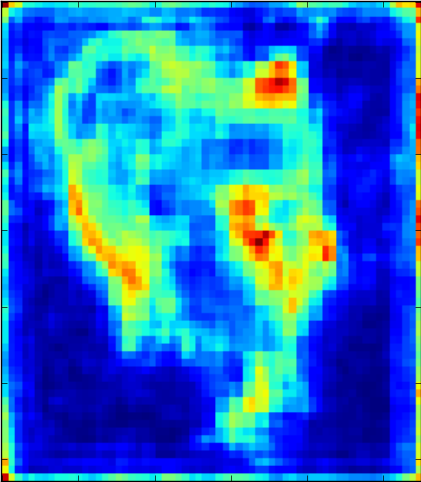} &
\includegraphics[width=.09\textwidth]{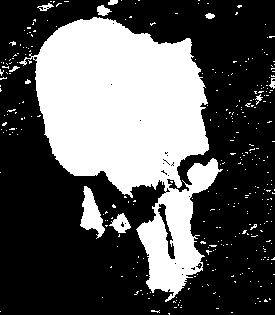} &
\includegraphics[width=.09\textwidth]{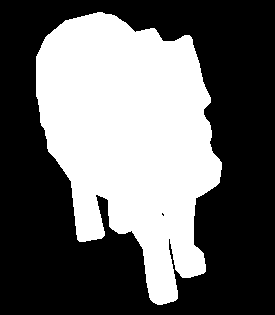} \\

\includegraphics[width=.09\textwidth]{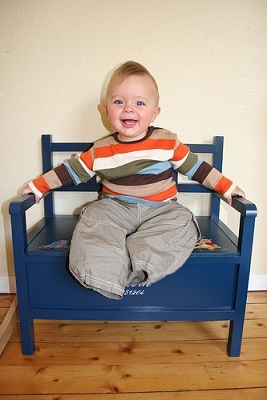} &
\includegraphics[width=.09\textwidth]{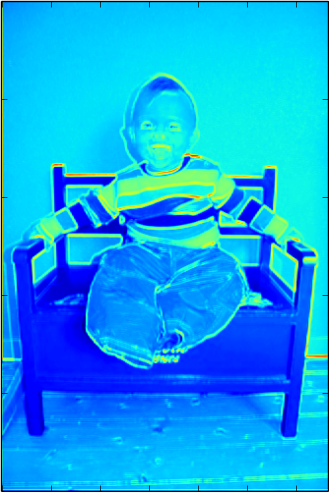} &
\includegraphics[width=.09\textwidth]{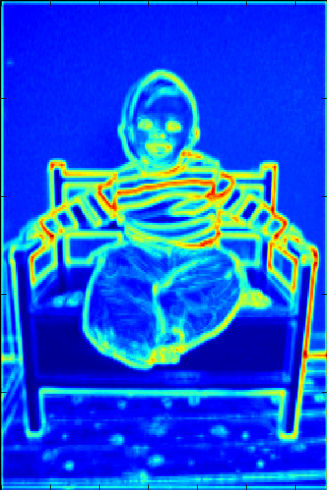} &
\includegraphics[width=.09\textwidth]{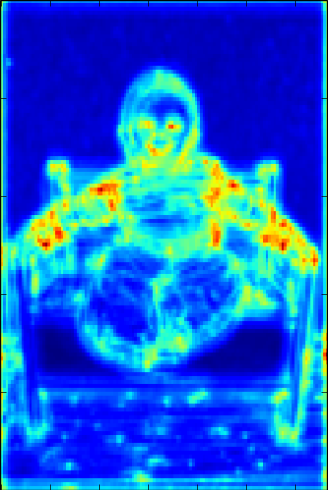} &
\includegraphics[width=.09\textwidth]{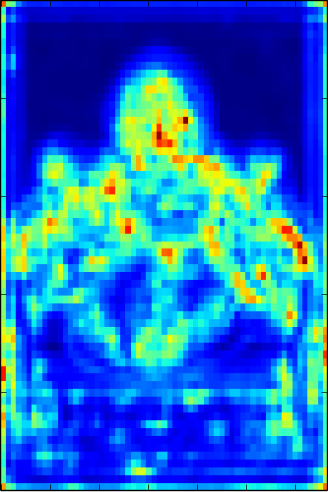} &
\includegraphics[width=.09\textwidth]{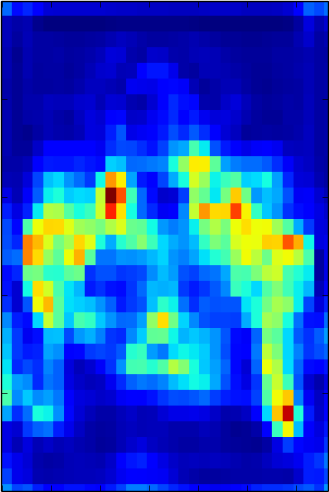} &
\includegraphics[width=.09\textwidth]{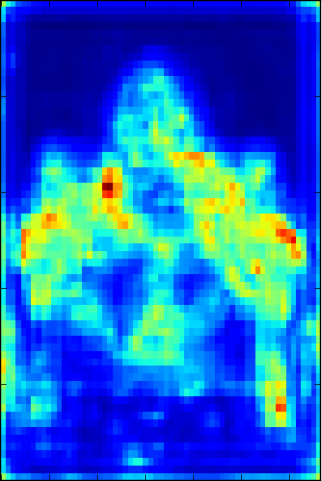} &
\includegraphics[width=.09\textwidth]{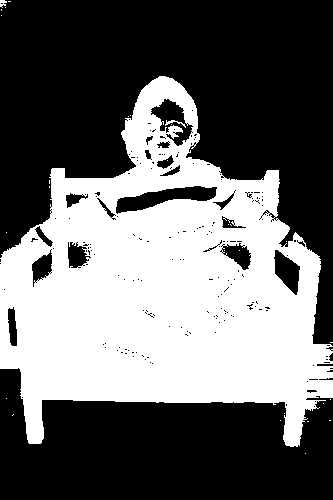} &
\includegraphics[width=.09\textwidth]{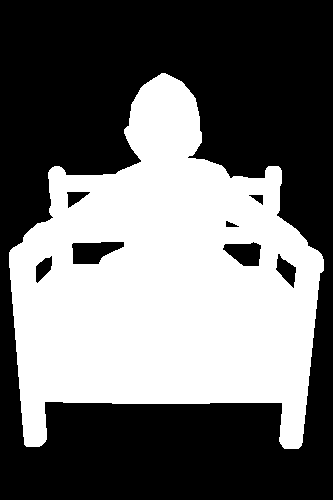} \\

\end{tabular}
\caption{{\bf Built-in foreground/background mask.} From left to right, there is the image, the activations of  \(1^{st}\), \(2^{nd}\),\(3^{rd}\) ,\(4^{th}\), and \(5^{th}\) Conv. layers, the results of our fusion strategy, and the final mask after CRF smoothing followed by G.T. Note that "Fusion" constitutes the unary potential of the dense CRF used to obtain "Our mask".}
\label{fig:fusion}
\end{figure}

Intuitively, we expect that a network trained for an object recognition task has learned to focus on the objects themselves, and their parts, rather than on background regions. In other words, it should produce high activation values on objects and their parts.
To evaluate this, we studied the activation of the different hidden layers of our initial network pre-trained on ImageNet. To this end, we forward each image through the network and visualize each activation by computing the mean over the channels after resizing the activation map to the input image size. Perhaps unsurprisingly, this lead to the following observations, illustrated in Fig.~\ref{fig:fusion}. The first two convolutional layers of the VGG network extract image edges. As we move deeper in the network, the convolutional layers extract higher-level features. In particular, the third convolutional layer fires up on prototypical object shapes.
The fourth layer indicates the location of complete objects, and the fifth one fires up on the most discriminative object parts~\cite{deepedge}.

Based on these observations, we propose to make use of the fourth and fifth layers to produce an initial foreground/background mask estimate. To this end, we first convert these two layers from 3D tensors (\(512\times W\times H\)) to 2D matrices (\(W\times H\)) via an average pooling operation over the 512 channels. We then fuse the two resulting matrices by simple elementwise sum, and scale the resulting values between 0 and 1. The resulting $W\times H$ map can be thought of as a pixelwise foreground probability. Fig.~\ref{fig:fusion} illustrates the results of this method on a few images from PASCAL VOC 2012. While the resulting scores indeed accurately indicate the location of the foreground objects, this initial mask remains noisy.

To overcome this, we therefore propose to exploit these foreground probabilities as unary potentials in a fully-connected CRF. Let ${\bf x} = \{x_i\}_{i=1}^{W\cdot H}$ be the set of random variables, where $x_i$ encodes the label of pixel $i$, i.e., either foreground or background. We encode the joint distribution over all pixels with a Gibbs energy of the form
\begin{equation}
E({\bf x} = {\bf X}) = -\sum_i \log P_f(x_i = X_i) + \sum_{i}\sum_{j>i}\theta_{ij}(x_i=X_i, x_j=X_j)\;,
\label{eq:gibbs}
\end{equation}
where $P_f(x_i=X_i)$ is the probability of pixel $i$ taking label assignment $X_i$, obtained directly from the foreground probability of our initial fusion strategy. Following~\cite{DenseCRF}, we define the pairwise term $\theta_{ij}$ as a contrast-sensitive Potts model using two Gaussian kernels encoding color similarity and spatial smoothness. This form lets us make use of the filtering-based mean-field strategy of~\cite{DenseCRF} to perform inference efficiently. Some resulting masks are shown in the last column of Fig.~\ref{fig:fusion}.

Note that our foreground/background masks can be thought of as a form of objectness measure. While objectness has been used previously for weakly-supervised semantic segmentation (MCG and BING in~\cite{fromimgtopixel}, and the generic objectness~\cite{objectness} in~\cite{whatsthepoint}), the benefits of our approach are twofold. First, we extract this information directly from the same network that will be used for semantic segmentation, which prevents us from having to rely on an external method. Second, as opposed to BING and MCG, we require neither object bounding boxes, nor object segments to train our method. While~\cite{objectness} predicts objectness after  training on a set of images, as shown in our experiments in the supplementary materials, our method yields much more accurate object localization than this technique.To further evidence the benefits of our approach, in supplementary material, we evaluate the masks obtained using the probabilities of~\cite{MCG} and~\cite{objectness} as unary potentials in the same dense CRF.

\subsection{Weakly-Supervised Learning}
\label{sec:learning}
We now introduce our learning algorithm for weakly-supervised semantic segmentation. We first introduce a simple loss based on image tags only, and then show how we can incorporate our foreground/background masks in our framework.

Intuitively, given image tags, one would like to encourage the image pixels to be labeled as one of the classes that are observed in the image, while preventing them to be assigned to unobserved classes. Note that this assumes that the tags cover all the classes depicted in the image. This assumption, however, is commonly employed in weakly-supervised semantic segmentation~\cite{whatsthepoint,MIL,fromimgtopixel}. Formally, given an input image $I$, let $\mathcal{L}$ be the set of classes that are present in the image (including background) and $\bar{\mathcal{L}}$ the set of classes that are absent. Furthermore, let us denote by $s_{i,j}^k(\theta)$ the score produced by our network with parameters $\theta$ for the pixel at location ($i,j$) and for class $k$, $0\leq k < N$. Note that, in general, we will omit the explicit dependency of the variables on the network parameters. Finally, let $S_{i,j}^k$ be the probability of class $k$ obtained after a softmax layer, i.e.,

\begin{equation}
S_{i,j}^k = \frac{\exp(s_{i,j}^k)}{\sum_{c=1}^N\exp(s_{i,j}^c)}\;.
\end{equation}

Encoding the above-mentioned intuition can then simply be achieved by designing a loss of the form
\begin{equation}
L_{weak} = -\frac{1}{|\mathcal{L}|}\sum_{k\in \mathcal{L}} \log{S^k} - \frac{1}{|\bar{\mathcal{L}}|}\sum_{k\in \bar{\mathcal{L}}}\log(1-S^k) \;,
\label{eq:loss_mil}
\end{equation}
where $S^k$ represents a candidate score for each class in the image. In short, the first term in Eq.~\ref{eq:loss_mil} expresses the fact that the present classes should be in the image, while the second term penalizes the pixels that have high probabilities for the absent classes. In practice, instead of computing $S^k$ as the maximum probability (as previously used in~\cite{MIL,whatsthepoint}) for class $k$ over all pixels in the image, we make use of the convex Log-Sum-Exp (LSE) approximation of the maximum (as previously used in~\cite{fromimgtopixel}), which can be written as
\begin{equation}
\tilde{S}^k = \frac{1}{r}\log\left[\frac{1}{|I|}\sum_{i,j \in I}\exp(rS_{i,j}^k)\right]\;,
\label{eq:LSE}
\end{equation}
where $|I|$ denotes the total number of pixels in the image and $r$ is a parameter allowing this function to behave in a range between the maximum and the average. In practice, following~\cite{fromimgtopixel}, we set $r$ to 5.

The loss in Eq.~\ref{eq:loss_mil} does not rely on any notion of foreground and background. As a consequence, minimizing it will typically yield poor object localization accuracy. To overcome this issue, we propose to make use of our built-in foreground/background mask introduced in Section~\ref{sec:fb}. Let $M_{i,j}$ denote the mask value at pixel $(i,j)$, i.e., $M_{i,j} = 1$ if pixel $(i,j)$ belongs to the foreground and 0 otherwise. We can then re-write our loss as
\begin{equation}
L_{mask} = -\frac{1}{|\mathcal{L}|-1}\sum_{k\in \mathcal{L}, \\ k \neq 0}\log(S_f^k) -\log(S^0) - \frac{1}{|\bar{\mathcal{L}}|.|I|}\sum_{i,j\in I,\;k\in \bar{\mathcal{L}}}\log(1-S_{i,j}^k)\;,
\label{eq:loss_mil_mask}
\end{equation}
where
\begin{equation}
S_f^k = \frac{1}{r}\log\left[\frac{1}{|M|}\sum_{i,j | M_{i,j}=1}\exp(rS_{i,j}^k)\right]\;,
\end{equation}
and
\begin{equation}
S^0 = \frac{1}{r}\log\left[\frac{1}{|\bar{M}|}\sum_{i,j | M_{i,j}=0}\exp(rS_{i,j}^0)\right]\;.
\end{equation}
where $|M|$ and $|\bar{M}|$ denote the number of foreground and background pixels, respectively, and $S_f^k$ computes an approximate maximum probability for the present class $k$ over all pixels in the foreground mask. Similarly, $S^0$ denotes an approximate maximum probability for the background class over all pixels outside the foreground mask. In short, the loss of Eq.~\ref{eq:loss_mil_mask} favors present classes to appear in the foreground mask, while pixels predicted as background should be assigned to the background class and no pixels should take on an absent label.

To learn the parameters of our network, we follow a standard back-propagation strategy to search for the parameters $\theta$ that minimize the loss in Eq. \ref{eq:loss_mask}. In particular, the network is fine-tuned using stochastic gradient descent (SGD) with momentum \(\mu\) to update the weights by a linear combination of the negative gradient and the previous weight update. At inference time, given the test image, the network performs a dense prediction. We optionally apply a fully connected CRF to smooth the segmentation using the default parameters of~\cite{DeepLab}.

\subsubsection{Remark.} Although our loss function performs well, an alternative to this loss function can be expressed as
\begin{equation}
L_{weak} = -\frac{1}{|I|}\sum_{i,j \in I}\log(S_{i,j}) - \frac{1}{|I|}\sum_{i,j\in I,\;  k\in \bar{\mathcal{L}}}\log(1-S_{i,j}^k)\;,
\label{eq:loss_weak}
\end{equation}
where $S_{i,j}$ represents the approximation of the maximum by using the LSE over the observed classes for each pixel as

\begin{equation}
S_{i,j} = \frac{1}{r}\log\left[\frac{1}{|\mathcal{L}|}\sum_{k \in \mathcal{L}}\exp(rS_{i,j}^k)\right]\;.
\label{eq:LSEremark}
\end{equation}

Such a formulation can also be extended to incorporate our mask, which yields
\begin{equation}
L_{mask} = -\frac{1}{|M|}\sum_{i,j | M_{i,j}=1}\log(S^f_{i,j}) -\frac{1}{|\bar{M}|}\sum_{i,j | M_{i,j}=0}\log(S_{i,j}^0) - \frac{1}{|I|}\sum_{i,j\in I,\;  k\in \bar{\mathcal{L}}}\log(1-S_{i,j}^k)\;,
\label{eq:loss_mask}
\end{equation}
where $S_{i,j}^0$ denotes the probability for the background class and
\begin{equation}
S^f_{i,j} = \frac{1}{r}\log\left[\frac{1}{|\mathcal{L}|-1}\sum_{k \in \mathcal{L}, \; k \neq 0}\exp(rS_{i,j}^k)\right]
\end{equation}
computes an approximate maximum probability over the observed foreground classes.

We found that, while the two approaches starting from the losses in Eq.~\ref{eq:loss_mil} and Eq.~\ref{eq:loss_weak} differ from each other, incorporating our masks in both of them using Eq.~\ref{eq:loss_mil_mask} and Eq.~\ref{eq:loss_mask} improves the segmentation quality considerably. Empirically, however, we found that this second formulation was slightly less effective than the one in Eq.~\ref{eq:loss_mil_mask}. This will be further discussed in the experiments.

\subsection{A Novel Weak Supervision: The \textit{CheckMask} Procedure}

The masks obtained with the approach introduced in Section~\ref{sec:fb} are not always perfect. This is due to the fact that the information obtained by fusing the activations of the fourth and fifth layers is noisy, and thus the solution found by inference in the CRF is not always the desired one. As a matter of fact, many other solutions also have a low energy (Eq.~\ref{eq:gibbs}). Rather than relying on a single mask prediction, we propose to generate multiple such predictions, and provide them to a user who decides which one is the best one.

The problem of generating several predictions in a given CRF is known as the $M$-best problem. Here, in particular, we are interested in generating solutions that all have low energy, but are diverse, and thus follow the approach of~\cite{diversembest}. In essence, this approach iteratively generates solutions, and, at each iteration, modifies the energy of Eq.~\ref{eq:gibbs} to encourage the next solution to be different from the ones generated previously. In practice, we make use of the Hamming distance as a diversity measure. This diversity measure can be encoded as an additional unary potential in Eq.~\ref{eq:gibbs}, and thus comes at virtually no additional cost in the inference procedure. For more details about the diverse $M$-best strategy, we refer the reader to~\cite{diversembest}.

\begin{figure}[t!]
\centering
\begin{tabular} {c c}
\includegraphics[width=.4\textwidth]{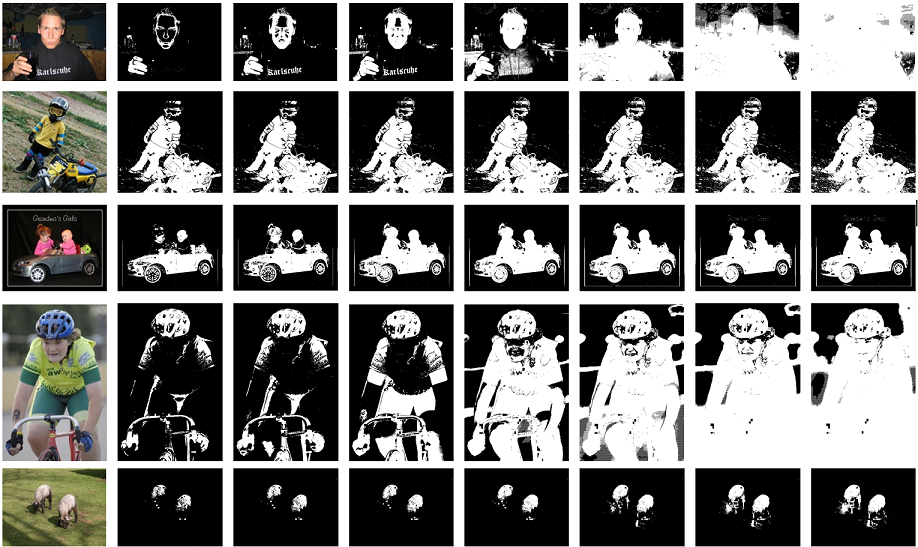} &
\includegraphics[width=.43\textwidth]{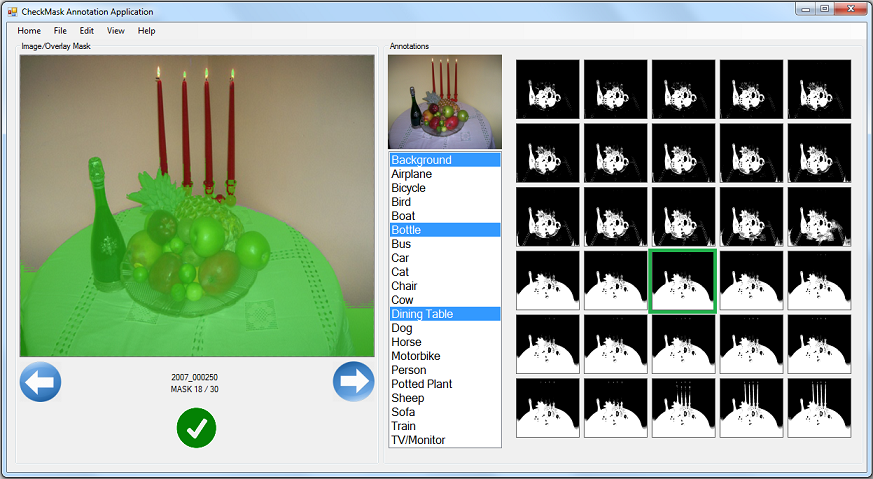}\\
(a) & (b) \\
\end{tabular}
\caption{(a) Mask candidates generated with our approach. From left to right, we show the input image, the \(1^{st}\), \(5^{th}\), \(10^{th}\), \(15^{th}\), \(20^{th}\), \(25^{th}\) and \(30^{th}\) solutions. (b) Our new level of supervision: The annotator selects a mask which he/she thinks contains all foreground object(s) and the minimum amount of background.}
\label{fig:checkmask}
\end{figure}

Ultimately, we generate several masks with this procedure, and ask the user to click on the one that best matches the input image. Such a selection can be achieved very quickly. In practice, we found that a user takes roughly 2-3 seconds per image to select the best mask. As a consequence, this new source of weak supervision remains very cheap, while, as evidenced by our experiments, allows us to achieve a significant improvement over our tags-only formulation (Fig.~\ref{fig:checkmask}).

\section{Experiments}
In this section, we first describe the datasets used for our experiments, and give some details about our learning and inference procedures. We then compare our approach to the state-of-the-art methods that use the same level of supervision as us. We provide an evaluation of our foreground/background masks in supplementary material.

\subsection{Datasets}
In our experiments, we first made use of the standard Pascal VOC 2012 dataset~\cite{pascal}, which serves as a benchmark in most weakly-supervised semantic segmentation papers~\cite{whatsthepoint,weaklyandsemi,MIL,fromimgtopixel,CCNN}. Similar to the dataset used in~\cite{whatsthepoint,CCNN,weaklyandsemi}, this dataset contains $N=21$ classes, and 10,582 training images (the VOC 2012 training set and the additional data annotated by~\cite{augmenteddata}), 1,449 validation images and 1,456 test images. The image tags were obtained from the pixel-level annotations by simply listing the classes observed in each image. As in~\cite{whatsthepoint,weaklyandsemi,fromimgtopixel,CCNN}, we report results on both the validation and the test set.

To further demonstrate the generality of our approach, we applied our method to a dataset that truly contains only image tags. To this end, we created a new training dataset from a subset of the MIRFLICKR-1M dataset~\cite{flickr}. In order to facilitate comparison, this subset was built using images containing the same classes as Pascal VOC 2012. In total it contains 7238 images, which were used for training purposes only. This new Flickr-based dataset does not provide any ground-truth pixel level annotations and, hence, the Pascal VOC validation set was used as test data. This training data will be made publicly available upon acceptance of the paper.

For both datasets, we report the mean intersection over union (mIOU), averaged over the 21 classes.

\subsection{Implementation details}
Our network architecture is detailed in Section~\ref{sec:architecture}. The parameters of this network were found by using stochastic gradient descent with a fixed learning rate of \(10^{-4}\) for the first 40k iterations, \(10^{-5}\) for the next 20k iterations, a momentum of \(0.9\), a weight decay of \(0.0005\), and mini-batches of size 1. Similar to recent weakly-supervised segmentation methods~\cite{fromimgtopixel,CCNN,MIL,whatsthepoint,weaklyandsemi}, the network weights were initialized with those of a network pre-trained for a 1000-way classification task on the ILSVRC 2012 dataset~\cite{ILSVRC}. Hence, for the last convolutional layer, we used the weights corresponding to the 20 classes shared by Pascal VOC and ILSVRC. For the background class, we initialized the weights with zero-mean Gaussian noise with a standard deviation of 0.1.
At inference time, given only the test image, the network generates a dense prediction as a complete semantic segmentation map. We used C++ and Python (Caffe framework~\cite{caffe}) for our implementation. As other methods~\cite{CCNN,weaklyandsemi}, we further optionally apply a dense CRF to refine this initial segmentation. To this end, we used the same CRF parameter values as these other approaches, i.e., the same as in~\cite{DeepLab}.

\subsection{Semantic Segmentation Results}

We now compare our approach with state-of-the-art baselines. We first present the results obtained with image tags only, and then those with additional weak supervision.
For the sake of completeness, in addition to the state-of-the-art baselines, we also report the results of our approach without using our foreground/background masks, i.e., by using Eq.~\ref{eq:loss_mil} as training loss. We also provide the segmentation results achieved by training a model using the losses introduced in Eq.~\ref{eq:loss_weak} and~\ref{eq:loss_mask}.
In the following, we will refer to our baseline as \emph{Ours (baseline)}, to our approach with tags only as \emph{Ours (tags)} and to our approach with additional weak supervision as \emph{Ours (CheckMask)}. We indicate the additional use of a dense CRF to further refine our results with \emph{+ CRF} after the method's name.

\subsubsection{Pascal VOC with image tags.}
In Table~\ref{tab:pascal_tags}, we compare our approach with our mask-free baseline and state-of-the-art methods on the task of semantic segmentation given only image tags during training. Note that our approach outperforms all the baselines by a large margin, whether we use CRF smoothing or not. Importantly, we outperform the methods based on an objectness prior~\cite{whatsthepoint,fromimgtopixel}, which clearly shows the benefits of using our built-in foreground/background masks instead of external objectness algorithms. The importance of our mask is further evidenced by the fact that we outperform our mask-free baseline by 13.8 mIOU points. Note that the best-performing baseline (MIL w/ILP)~\cite{fromimgtopixel} uses a large amount of additional images (roughly 700K) from the ILSVRC2013 dataset to boost the accuracy of the basic MIL method. Note that we still outperform this baseline, even without using any such additional data.

\begin{table}[t!]
\centering
\tiny
\caption{Per class IOU on the PASCAL VOC 2012 validation set for methods trained using image tags. }
\label{tab:pascal_tags}
\scalebox{0.80}
{
\begin{tabular}{| l|c c c c c c c c c c c c c c c c c c c c c|c |}
\hline
Method & \rotatebox[origin=c]{90}{bg} & \rotatebox[origin=c]{90}{aero} & \rotatebox[origin=c]{90}{bike} & \rotatebox[origin=c]{90}{bird} & \rotatebox[origin=c]{90}{boat} & \rotatebox[origin=c]{90}{bottle} & \rotatebox[origin=c]{90}{bus} & \rotatebox[origin=c]{90}{car} & \rotatebox[origin=c]{90}{cat} & \rotatebox[origin=c]{90}{chair} & \rotatebox[origin=c]{90}{cow} & \rotatebox[origin=c]{90}{table} & \rotatebox[origin=c]{90}{dog} & \rotatebox[origin=c]{90}{horse} & \rotatebox[origin=c]{90}{mbike} & \rotatebox[origin=c]{90}{person} & \rotatebox[origin=c]{90}{plant} & \rotatebox[origin=c]{90}{sheep}& \rotatebox[origin=c]{90}{sofa} & \rotatebox[origin=c]{90}{train} & \rotatebox[origin=c]{90}{tv} & \rotatebox[origin=c]{0}{mIOU} \\
 \hline
Ours (baseline),Eq.~\ref{eq:loss_weak} & 65.4 &   19.1 &    14.1 &    19.5 &   18.8 &    25.5 &    36.9 &    34.6  &  43.0  &  9.0 &   11.9  &  14.6   & 33.7  &  21.1 &   33.0   & 35.3 &   25.2 &	15.2 &    10.3 &    30.2 &    29.8 & 26.0 \\
Ours (baseline) & 59.2 & 25.2 & 14.6 & 21.9 & 19.0 & 28.6 & 49.5 & 41.9 & 42.7 & 10.1 & 32.9 & 25.6 & 36.8 & 29.8 & 34.5 & 32.5 & 24.6 & 31.2 & 21.6 & 39.3 & 29.0 & 31.0 \\
\hline
MIL(Tag)~\cite{fromimgtopixel} & 37.0  & 10.4  & 12.4 &  10.8 & 5.3 & 5.7 & 25.2 & 21.1 & 25.15 & 4.8 & 21.5 & 8.6 & 29.1 & 25.1 & 23.6 & 25.5 & 12.0 & 28.4 & 8.9 & 22.0 & 11.6 & 17.8\\

MIL(Tag) w/ILP~\cite{fromimgtopixel} & 73.2 & 25.4 & 18.2 & 22.7 & 21.5 & 28.6 & 39.5 & 44.7 & 46.6 & 11.9 & 40.4 & 11.8 & 45.6 & 40.1 & 35.5 & 35.2 & 20.8 & 41.7 & 17.0 & 34.7 & 30.4 & 32.6\\

MIL(Tag) w/ILP+sspxl~\cite{fromimgtopixel} & 77.2 & 37.3 & 18.4 & 25.4 & 28.2 & 31.9 & 41.6 & 48.1 & 50.7 & 12.7 & 45.7 & 14.6 & 50.9 & 44.1 & 39.2 & 37.9 & 28.3 & 44.0 & 19.6 & 37.6 & 35.0 & 36.6\\

What's the point(Tag) W/Obj~\cite{whatsthepoint}& 78.8& 41.6& 19.8& 38.7& 33.0& 17.2& 33.8& 38.8& 45.0& 10.4& 35.2& 12.6& 42.3& 34.3& 33.2& 22.7& 18.6& 40.1& 14.9& 37.7& 28.1& 32.2\\

EM-Fixed(Tag)+CRF~\cite{weaklyandsemi} & - & - & - & - & - & - & - & - & - & - & - & - & - & - & - & - & - & - & - & - & - & 20.8 \\

EM-Adapt(Tag)+CRF~\cite{weaklyandsemi} & - & - & - & - & - & - & - & - & - & - & - & - & - & - & - & - & - & - & - & - & - & 38.2 \\

CCNN(Tag)~\cite{CCNN} & 66.3 & 24.6 & 17.2 & 24.3 & 19.5 & 34.4 & 45.6 & 44.3 & 44.7 & 14.4 & 33.8 & 21.4 & 40.8 & 31.6 & 42.8 & 39.1 & 28.8 & 33.2 & 21.5 & 37.4 & 34.4 & 33.3\\

CCNN(Tag)+CRF~\cite{CCNN} & 68.5 & 25.5 & 18.0 & 25.4 & 20.2 & 36.3 & 46.8 & 47.1 & 48.0 & 15.8 & 37.9 & 21.0 & 44.5 & 34.5 & 46.2 & 40.7 & 30.4 & 36.3 & 22.2 & 38.8 & 36.9 & 35.3\\
\hline

Ours (Tags),Eq.~\ref{eq:loss_mask} & 79.7 & 56.4 & 19.2 & 54.0 & 40.9 & 44.7 & 62.9 & 49.9 & 56.2 & 12.2 & 44.5 & 33.7 & 55.0 & 44.1 & 52.2 & 46.2 & 32.3 & 50.0 & 26.7 & 52.9 & 26.5 & 44.8\\
Ours (Tags)+CRF,Eq.~\ref{eq:loss_mask} & 81.0 & 57.6 & 19.7 & 45.8 & 41.5 & 46.4 & 62.5 & 53.2 & 59.3 & 14.0 & 44.4 & 38.0 & 55.4 & 46.5 & 54.2 & 49.3 & 34.2 & 50.7 & 27.1 & 55.3 & 25.9 & 45.8\\
 \hline

Ours (Tags) & 79.7 & 56.2 & 19.1 & 53.8 & 41.3 & 44.6 & 62.8 & 50.1 & 56.1 & 12.1 & 44.8 & 33.9 & 54.9 & 44.3 & 52.3 & 46.1 & 33.1 & 49.5 & 26.9 & 52.7 & 26.7 & 44.8 \\
Ours (Tags)+CRF & 79.2 & 60.1 & 20.4 & 50.7 & 41.2 & 46.3 & 62.6 & 49.2 & 62.3 & 13.3 & 49.7 & 38.1 & 58.4 & 49.0 & 57.0 & 48.2 & 27.8 & 55.1 & 29.6 & 54.6 & 26.6 & 46.6\\
  \hline
\end{tabular}
}
\end{table}

\subsubsection{Pascal VOC with additional weak supervision.}
We then evaluate our approach on Pascal VOC with our additional CheckMask weak supervision procedure. While no other approaches have used this same kind of weak supervision, we report the results of methods that have used additional weak supervision of a similar cost to compute. In particular, these includes the point supervision of~\cite{whatsthepoint}, the random crops of~\cite{weaklyandsemi}, the size information of~\cite{CCNN} and the MCG segments of~\cite{fromimgtopixel,learnTosegwithimagelevelanno}. The results of this comparison are provided in Table~\ref{tab:pascal_weak}. Note that our CheckMask procedure yields an improvement of 4.2 mIOU point (and 4.9 mIOU point when a CRF is applied) over our tag-only approach. More importantly, our approach outperforms the baselines by a large margin. Note that other approaches have proposed to rely on labeled bounding boxes, which require a user to provide a bounding box for each individual foreground object in an image and to associate a label to each such bounding box. While this procedure is clearly more costly than ours, we achieve similar accuracy to these baselines (52.5\% for~\cite{weaklyandsemi} when using labeled bounding boxes and 54.1\% for~\cite{weaklyandsemi} when using labeled bounding boxes in an EM process vs 51.49\% for our approach). We believe that this further evidences the benefits of our approach.
We also report the results on the test set of Pascal VOC 2012 and compare our method with other baselines (see Table~\ref{tab:pascal_test}).

\begin{table}[t!]
\centering
\scriptsize
\caption{Per class IOU on the PASCAL VOC 2012 validation set using additional supervision during training.}
\label{tab:pascal_weak}
\scalebox{0.72}
{
\begin{tabular}{|l|c c c c c c c c c c c c c c c c c c c c c|c |}
\hline
Method: Additional Supervision & \rotatebox[origin=c]{90}{bg} & \rotatebox[origin=c]{90}{aero} & \rotatebox[origin=c]{90}{bike} & \rotatebox[origin=c]{90}{bird} & \rotatebox[origin=c]{90}{boat} & \rotatebox[origin=c]{90}{bottle} & \rotatebox[origin=c]{90}{bus} & \rotatebox[origin=c]{90}{car} & \rotatebox[origin=c]{90}{cat} & \rotatebox[origin=c]{90}{chair} & \rotatebox[origin=c]{90}{cow} & \rotatebox[origin=c]{90}{table} & \rotatebox[origin=c]{90}{dog} & \rotatebox[origin=c]{90}{horse} & \rotatebox[origin=c]{90}{mbike} & \rotatebox[origin=c]{90}{person} & \rotatebox[origin=c]{90}{plant} & \rotatebox[origin=c]{90}{sheep}& \rotatebox[origin=c]{90}{sofa} & \rotatebox[origin=c]{90}{train} & \rotatebox[origin=c]{90}{tv} & \rotatebox[origin=c]{0}{mIOU} \\
 \hline
 \cite{fromimgtopixel}: MIL(Tag) w/ILP+bbox & 78.6 & 46.9 & 18.6 & 27.9 & 30.7 & 38.4 & 44.0 & 49.6 & 49.8 & 11.6 & 44.7 & 14.6 & 50.4 & 44.7 & 40.8 & 38.5 & 26.0 & 45.0 & 20.5 & 36.9 & 34.8 & 37.8\\

\cite{fromimgtopixel}: MIL(Tag) w/ILP+seg & 79.6 & 50.2 & 21.6 & 40.6 & 34.9 & 40.5 & 45.9 & 51.5 & 60.6 & 12.6 & 51.2 & 11.6 & 56.8 & 52.9 & 44.8 & 42.7 & 31.2 & 55.4 & 21.5 & 38.8 & 36.9 & 42.0 \\

\cite{learnTosegwithimagelevelanno}: SN-B+MCG seg & 80.7 & 54.6 & 10.7 & 55.6 & 37.5 & 51.8 & 46.3 & 42.6 & 48.0 & 16.0 & 46.3 & 10.0 & 54.6 & 45.9 & 47.5 & 34.4 & 24.5 & 53.7 & 23.0 & 47.8 & 48.6 & 41.9 \\
\cite{STC}: STC+Additional train data  & 84.5 & 68.0 &19.5& 60.5& 42.5& 44.8 &68.4 & 64.0 & 64.8 & 14.5 & 52.0 & 22.8 & 58.0 & 55.3 & 57.8 & 60.5 & 40.6  & 56.7 & 23.0  & 57.1 & 31.2 & 49.8
\\
\cite{whatsthepoint}: Objectness & 78.8 & 41.6 & 19.8 & 38.7 & 33.0 & 17.2 & 33.8 & 38.8 & 45.0 & 10.4 & 35.2 & 12.6 & 42.3 & 34.3 & 33.2  &22.7 & 18.6 & 40.1 & 14.9 & 37.7 & 28.1 & 32.2\\
\cite{whatsthepoint}: 1Point & 56.2 & 24.5 & 16.1 & 21.5 & 20.0 & 30.8 & 53.0 & 34.2 & 53.0 & 7.9 & 41.4 & 41.6 & 42.7 & 40.1 & 42.0 & 45.9 & 24.0 & 37.5 & 28.5  &45.6 & 29.8 & 35.1\\
\cite{whatsthepoint}: Objectness+1Point & 77.9 & 48.6 & 22.6 & 36.6 & 36.9 & 37.4 & 57.4 & 50.4 & 50.9 & 13.7 & 40.0 & 40.9 & 49.5 & 38.1 & 51.1 & 46.9 & 31.2 & 48.2 & 27.8 & 48.9 & 44.9 & 42.7\\
\cite{whatsthepoint}: Objectness+AllPoints & 78.5 & 48.5 & 21.3 & 39.5 & 37.9  &37.7 & 49.5 & 45.3 & 52.5 & 17.0 & 42.7  &39.9 & 46.8 & 44.0 & 51.0 & 50.6 & 22.0 & 46.6 & 28.9 & 52.3 & 44.3 & 42.7\\

\cite{whatsthepoint}: Objectness+1Point(GT) & 79.6 & 49.4 & 22.9 & 38.6 & 40.9 & 45.8 & 60.4 & 60.9 & 55.5 & 17.7 & 37.8  &41.0  &54.1 & 41.7 & 54.9 & 56.9  &32.2 & 51.1 & 26.4 & 54.5  &45.3 & 46.1\\
\cite{CCNN}: Random Crops & - & - & - & - & - & - & - & - & - & - & - & - & - & - & - & - & - & - & - & - & - & 34.4 \\
\cite{CCNN}: Random Crops+CRF & - & - & - & - & - & - & - & - & - & - & - & - & - & - & - & - & - & - & - & - & - & 36.4 \\
\cite{CCNN}: Size Info. & - & - & - & - & - & - & - & - & - & - & - & - & - & - & - & - & - & - & - & - & - & 40.5\\
\cite{CCNN}: Size Info.+CRF & - & - & - & - & - & - & - & - & - & - & - & - & - & - & - & - & - & - & - & - & - &  42.4\\
\hline
Ours (CheckMask),Eq.~\ref{eq:loss_mask} &  85.2 & 65.2 & 21.4 & 49.7 & 50.2 & 48.5 & 67.3 & 63.2 & 58.5 & 16.1 & 43.3  &    36.1 & 56.1 & 45.4 & 53.7 & 52.3 & 36.5 & 47.1 & 27.0 & 61.9 & 37.2 & 48.7\\
Ours (CheckMask)+CRF,Eq.~\ref{eq:loss_mask} & 86.3 & 70.4 & 22.1 & 48.9 & 53.4 & 49.8 & 71.0 & 66.0 & 61.5 & 17.7 & 46.4 & 38.3 & 59.9 & 49.1 & 57.4 & 55.8 & 38.5 & 50.6 & 28.4 & 64.3 & 36.0 & 51.0\\
Ours (CheckMask) & 85.2 & 65.2 & 21.1 & 52.1 & 49.9 & 48.6 & 67.3 & 63.6 & 59.7 & 15.5 & 43.2 & 36.6 & 56.9 & 45.9 & 53.7 & 52.6 & 37.0 & 48.2 & 27.6 & 61.7 & 36.9 & 49.0\\
Ours (CheckMask)+CRF & 86.4 & 70.1 & 21.7 & 53.1 & 52.5 & 50.7 & 70.9 & 66.6 & 63.2 & 16.9 & 45.8 & 39.1 & 61.1 & 50.0 & 56.8 & 56.2 & 40.0 & 51.9 & 29.3 & 63.1 & 05.9 & 51.5  \\
 \hline
 \end{tabular}
 }
\end{table}
\begin{table}[t!]

\tiny
\caption{Per class IOU on the PASCAL VOC 2012 test set. }
\label{tab:pascal_test}
\scalebox{0.8}
{
\begin{tabular}{| l|c c c c c c c c c c c c c c c c c c c c c|c |}
\hline
Method & \rotatebox[origin=c]{90}{bg} & \rotatebox[origin=c]{90}{aero} & \rotatebox[origin=c]{90}{bike} & \rotatebox[origin=c]{90}{bird} & \rotatebox[origin=c]{90}{boat} & \rotatebox[origin=c]{90}{bottle} & \rotatebox[origin=c]{90}{bus} & \rotatebox[origin=c]{90}{car} & \rotatebox[origin=c]{90}{cat} & \rotatebox[origin=c]{90}{chair} & \rotatebox[origin=c]{90}{cow} & \rotatebox[origin=c]{90}{table} & \rotatebox[origin=c]{90}{dog} & \rotatebox[origin=c]{90}{horse} & \rotatebox[origin=c]{90}{mbike} & \rotatebox[origin=c]{90}{person} & \rotatebox[origin=c]{90}{plant} & \rotatebox[origin=c]{90}{sheep}& \rotatebox[origin=c]{90}{sofa} & \rotatebox[origin=c]{90}{train} & \rotatebox[origin=c]{90}{tv} & \rotatebox[origin=c]{0}{mIOU} \\
 \hline
CCNN (tags)~\cite{CCNN} & - & 24.2 & 19.9 & 26.3 & 18.6 & 38.1 &  51.7 &  42.9 &  48.2 &  15.6  & 37.2  & 18.3 &  43.0  & 38.2  & 52.2  & 40.0  & 33.8 - & 36.0  & 21.6  & 33.4 &  38.3  & 35.6\\
CCNN (tags)+size~\cite{CCNN} & - &  36.7 & 23.6 & 47.1 & 30.2 & 40.6 & 59.5 & 54.3 & 51.9 & 15.9 & 43.3 & 34.8 & 48.2 & 42.5 & 59.2 & 43.1 & 35.5 & 45.2 & 31.4 & 46.2 & 42.2 & 43.3\\
CCNN (tags)+size+CRF~\cite{CCNN} &  & 42.3 & 24.5 & 56.0 & 30.6 & 39.0 & 58.8 & 52.7 & 54.8 & 14.6 & 48.4 & 34.2 & 52.7 & 46.9 & 61.1 & 44.8 & 37.4 & 48.8 & 30.6 & 47.7 & 41.7 & 45.1\\

MIL-FCN~\cite{fromimgtopixel} & - & - & - & - & - & - & - & - & - & - & - & - & - & - & - & - & - & - & - & - & - & 25.7\\
MIL-sppxl~\cite{fromimgtopixel} & 74.7 & 38.8 & 19.8 & 27.5 & 21.7 & 32.8 & 40.0 & 50.1 & 47.1 & 7.2 & 44.8 & 15.8 & 49.4 & 47.3 & 36.6 & 36.4 & 24.3 & 44.5 & 21.0 & 31.5 & 41.3 & 35.8\\
MIL-obj~\cite{fromimgtopixel} & 76.2 & 42.8 & 20.9 & 29.6 & 25.9 & 38.5 & 40.6 & 51.7 & 49.0 & 9.1 & 43.5 & 16.2 & 50.1 & 46.0 & 35.8 & 38.0 & 22.1 & 44.5 & 22.4 & 30.8 & 43.0 & 37.0\\
MIL-seg~\cite{fromimgtopixel} & 78.7 & 48.0 & 21.2 & 31.1 & 28.4 & 35.1 & 51.4 & 55.5 & 52.8 & 7.8 & 56.2 & 19.9 & 53.8 & 50.3 & 40.0 & 38.6 & 27.8 & 51.8 & 24.7 & 33.3 & 46.3 & 40.6\\
WhatsThePnt+Obj+1Point~\cite{whatsthepoint} & 80.6 & 50.2 & 23.9 & 38.4 & 33.1 & 38.5 & 52.0 & 50.9 & 55.4 & 18.3 & 38.2 & 37.7 & 51.0 & 46.1 &  54.7  & 43.2  & 35.4  & 45.1  & 33.0  & 49.6  & 40.0 &  43.6\\
EM-Adapt+CRF~\cite{weaklyandsemi} & 76.3 & 37.1 & 21.9 & 41.6 & 26.1 & 38.5 & 50.8 & 44.9 & 48.9 & 16.7 & 40.8 & 29.4 & 47.1 & 45.8 & 54.8 & 28.2 & 30.0 & 44.0 & 29.2 & 34.3 & 46.0 & 39.6\\
SN-B+MCG seg~\cite{learnTosegwithimagelevelanno} & 82.1 & 53.6 & 12.4 & 53.5 & 29.5 & 41.6 & 46.9 & 46.3 & 50.3 & 16.8 & 48.7 & 17.2 & 60.6 & 51.8 & 61.7 & 36.4 & 25.2 & 58.3 & 19.3 & 48.5 & 45.5 & 43.2 \\

\hline
Ours (Tags)  & 80.6 & 54.7 & 22.0 & 63.2 & 34.0 & 44.3 & 64.5 & 49.4 & 53.0 & 12.5 & 45.6 & 38.8 & 53.9 & 45.0 & 61.6 & 42.5 & 40.3 & 51.3 & 31.0 & 42.5 & 32.0 & 45.8\\
Ours (Tags)+CRF & 80.3 & 57.5 & 24.1 & 66.9 & 31.7 & 43.0 & 67.5 & 48.6 & 56.7 & 12.6 & 50.9 & 42.6 & 59.4 & 52.9 & 65.0 & 44.8 & 41.3 & 51.1 & 33.7 & 44.4 & 33.2 & 48.0\\
Ours (CheckMask)  &	86.1&	61.8&	24.9&	60.8&	42.8&	49.7&	68.6&	61.7&	56.3&	15.5&	44.2&	41.0&	56.1&	47.2&	64.1&	50.2&	40.9&	48.1&	33.1&	55.6&	41.4&	50.0\\

Ours (CheckMask)+CRF  &	87.4&	65.7&	26.0&	64.2&	43.7&	53.2&	72.6&	63.6&	59.5&	17.1&	48.0&	43.7&	61.2&	52.0&	69.3&	54.8&	43.0&	50.3&	34.6&	59.2&	42.0&	52.9\\

  \hline
\end{tabular}
}

\end{table}

\subsubsection{
Flickr (MIRFLICKR-1M) with image tags and additional weak supervision.}
We now evaluate our method by training it using our new dataset containing a subset of the MIRFLICKR-1M images~\cite{flickr}. Since no other results have been reported on this dataset, we also computed the results of CCNN~\cite{CCNN} whose code is publicly available, and which has shown to yield good accuracy in the previous experiments. In Table~\ref{tab:flickr}, we compare the results of our approach with this baseline when trained using tags only, and, as mentioned before, tested on the Pascal VOC 2012 validation dataset, since no ground-truth pixel level annotations are available in Flickr. Note that our approach significantly outperforms both our mask-free baseline and the CCNN by a large margin. It is worth mentioning that this dataset contains three rare classes, Chair, Dining Table, and the Sofa which have 1.1\%, 0.5\%, and 1.3\% of the whole dataset respectively. Although these classes have a negligible contribution in constructing this dataset, our approach performs well in comparison to CCNN in segmenting these classes (17.0\% vs 10.7\%, 31.2\% vs 0\%, and 16.8\% vs 0\% in these classes respectively).

\begin{figure}[!t]
\centering
\tiny
\begin{tabular}{ccccccc}
 Image & baseline & Tags & CheckMask & Tags& CheckMask&GT  \\
 & Pascal & Pascal & Pascal &Flickr & Flickr &  \\
\includegraphics[width=.12\textwidth]{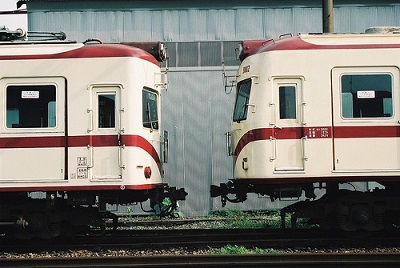} &
\includegraphics[width=.12\textwidth]{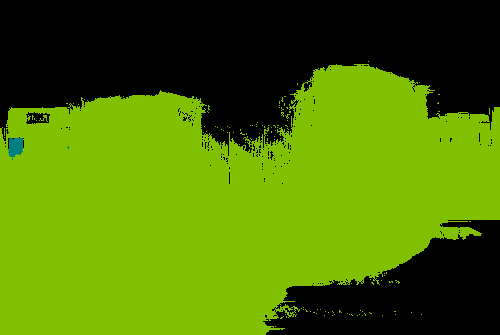} &
\includegraphics[width=.12\textwidth]{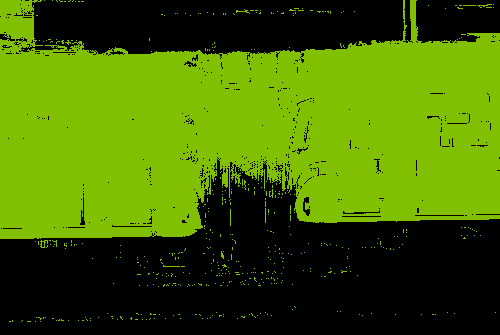} &
\includegraphics[width=.12\textwidth]{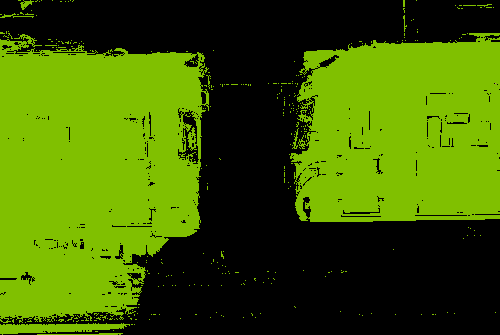} &
\includegraphics[width=.12\textwidth]{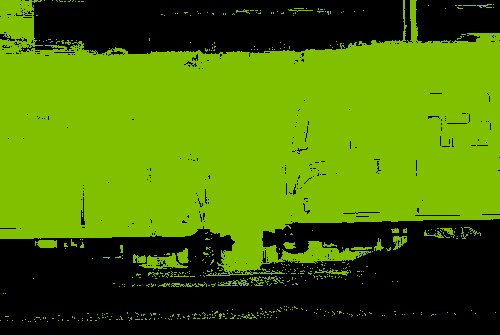} &
\includegraphics[width=.12\textwidth]{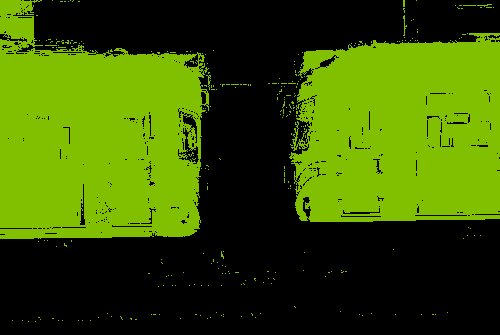} &
\includegraphics[width=.12\textwidth]{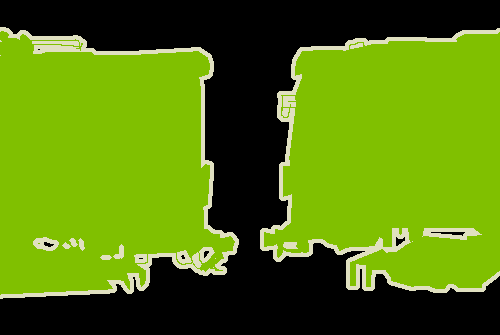} \\

\includegraphics[width=.12\textwidth]{2007_000783.jpg} &
\includegraphics[width=.12\textwidth]{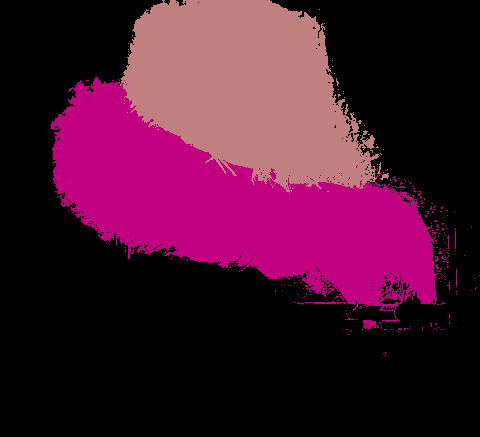} &
\includegraphics[width=.12\textwidth]{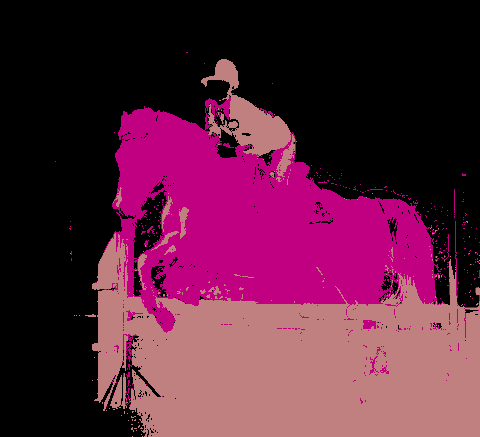} &
\includegraphics[width=.12\textwidth]{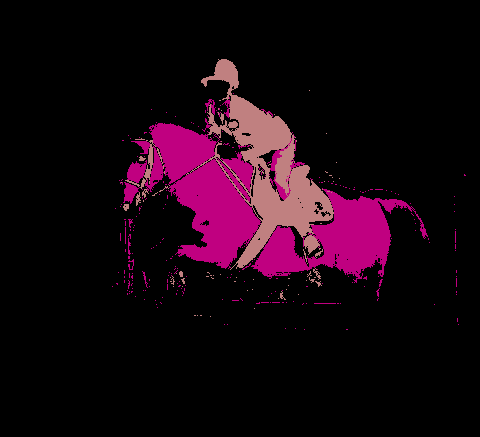} &
\includegraphics[width=.12\textwidth]{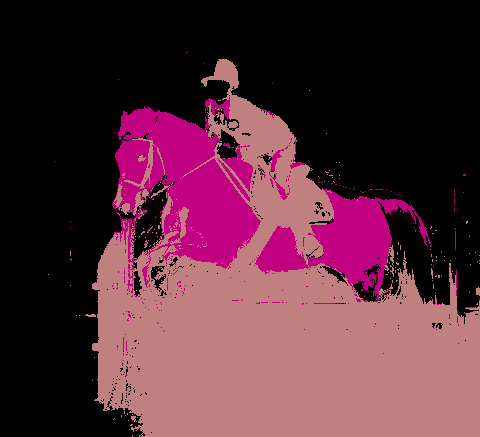} &
\includegraphics[width=.12\textwidth]{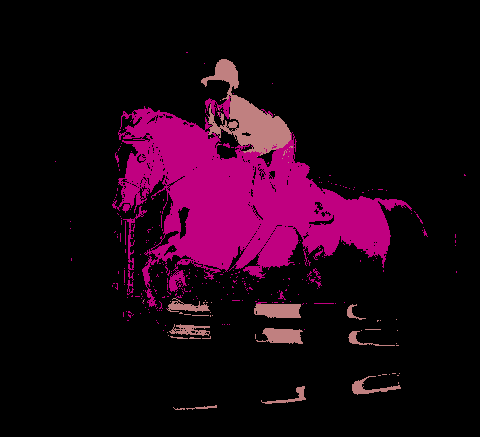} &
\includegraphics[width=.12\textwidth]{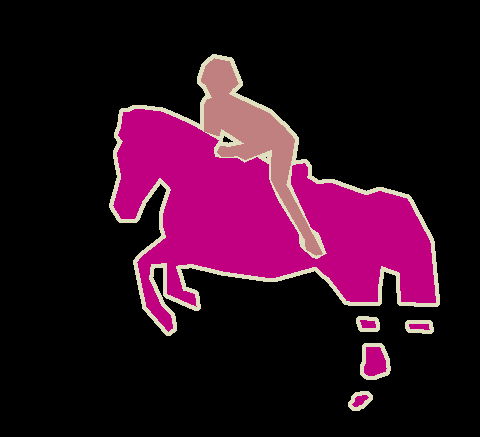} \\

\includegraphics[width=.12\textwidth]{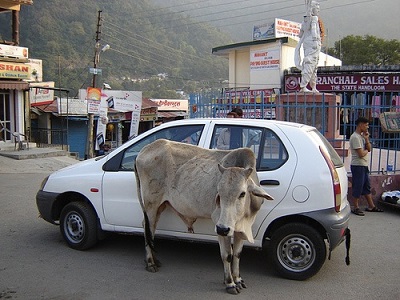} &
\includegraphics[width=.12\textwidth]{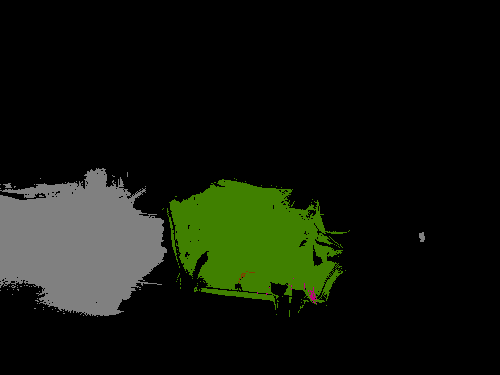} &
\includegraphics[width=.12\textwidth]{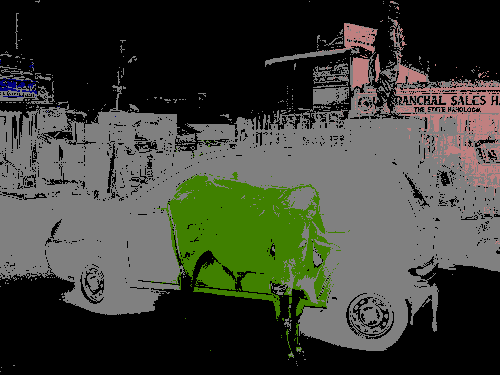} &
\includegraphics[width=.12\textwidth]{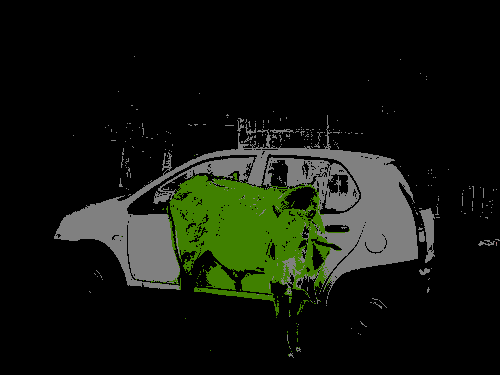} &
\includegraphics[width=.12\textwidth]{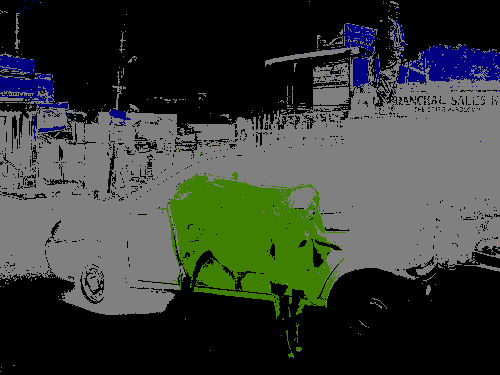} &
\includegraphics[width=.12\textwidth]{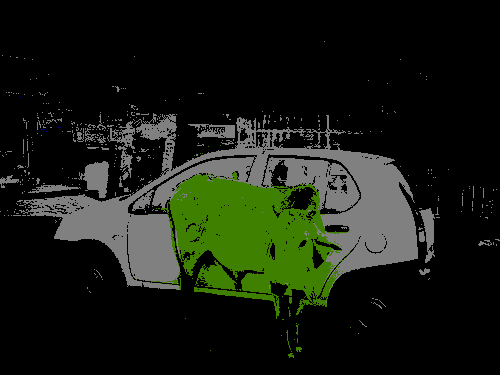} &
\includegraphics[width=.12\textwidth]{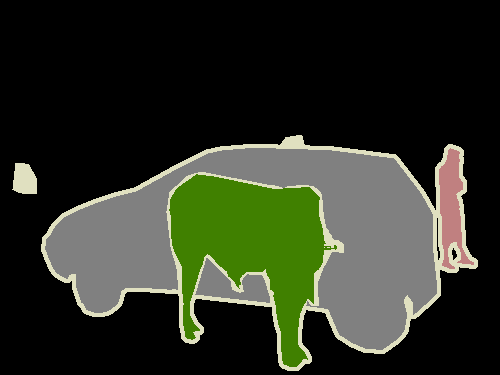} \\

\includegraphics[width=.12\textwidth]{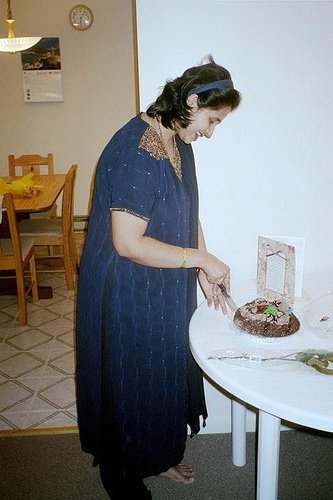} &
\includegraphics[width=.12\textwidth]{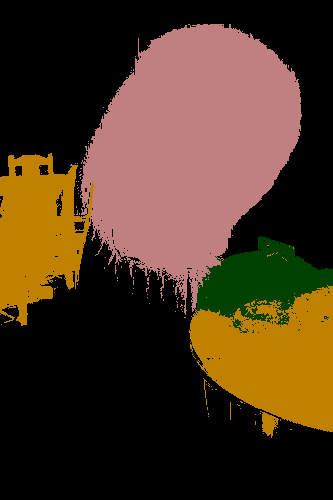} &
\includegraphics[width=.12\textwidth]{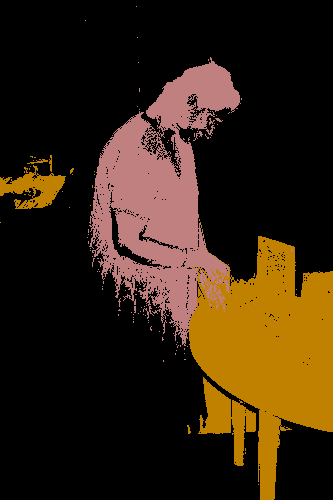} &
\includegraphics[width=.12\textwidth]{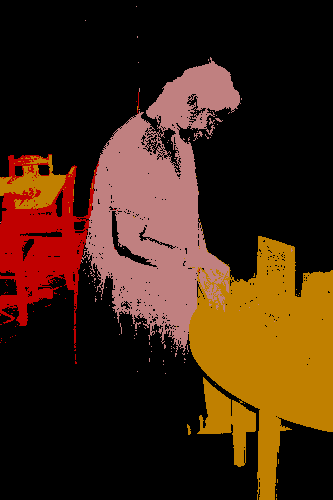} &
\includegraphics[width=.12\textwidth]{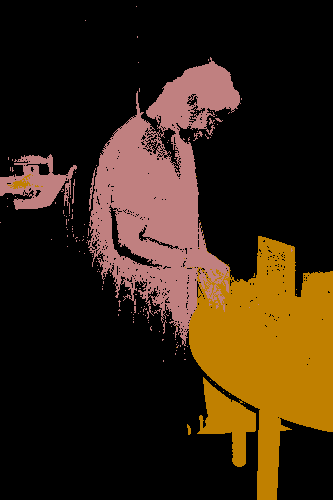} &
\includegraphics[width=.12\textwidth]{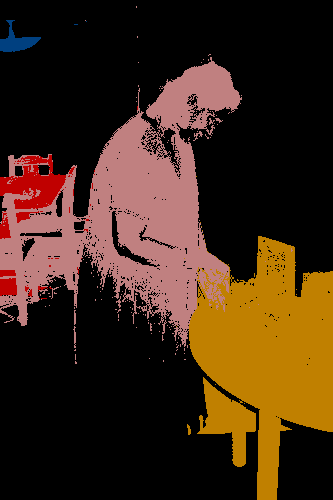} &
\includegraphics[width=.12\textwidth]{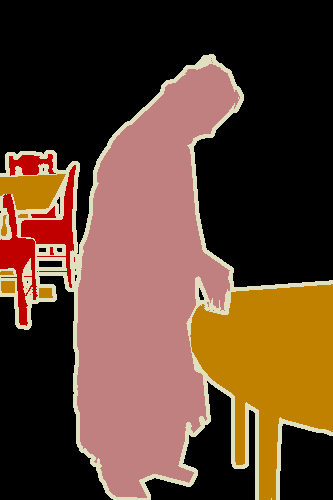} \\

\includegraphics[width=.12\textwidth]{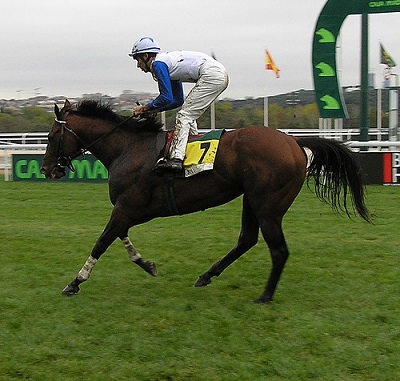} &
\includegraphics[width=.12\textwidth]{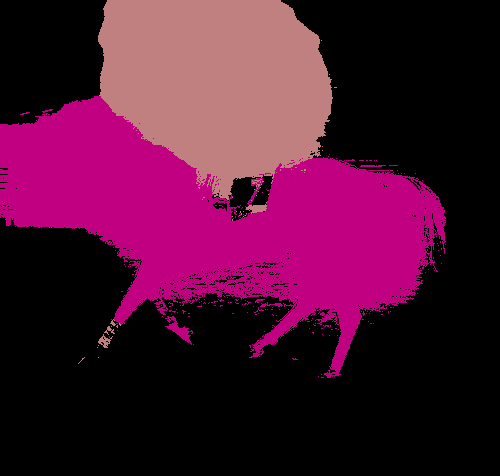} &
\includegraphics[width=.12\textwidth]{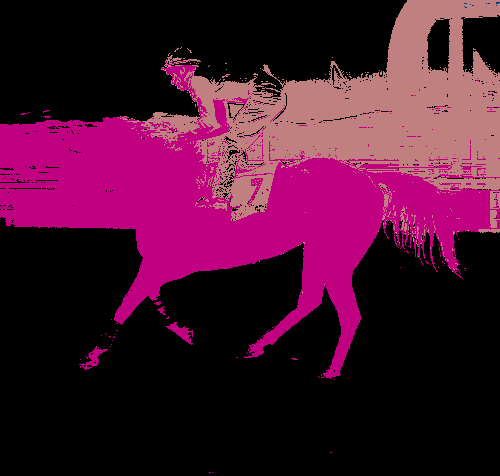} &
\includegraphics[width=.12\textwidth]{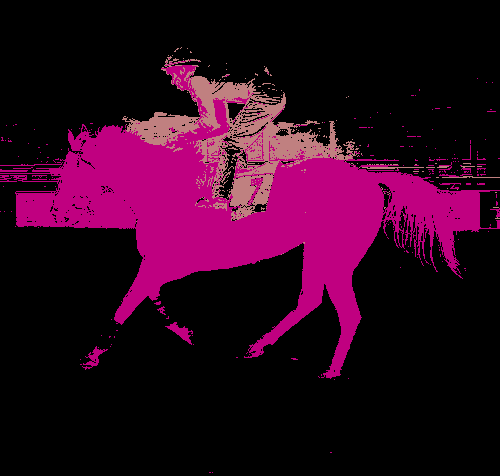} &
\includegraphics[width=.12\textwidth]{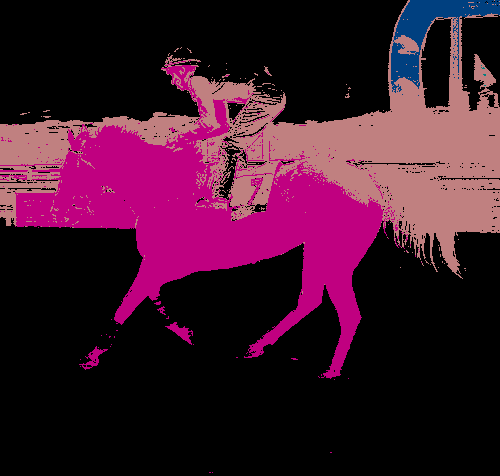} &
\includegraphics[width=.12\textwidth]{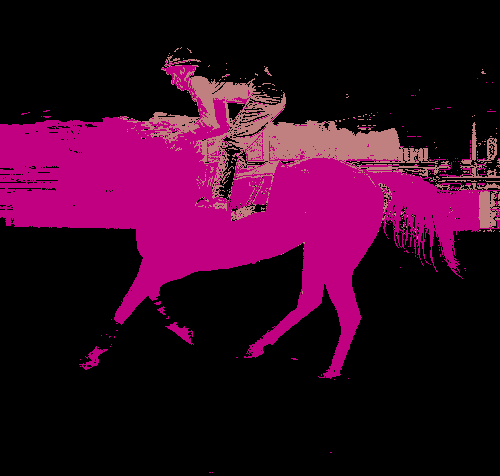} &
\includegraphics[width=.12\textwidth]{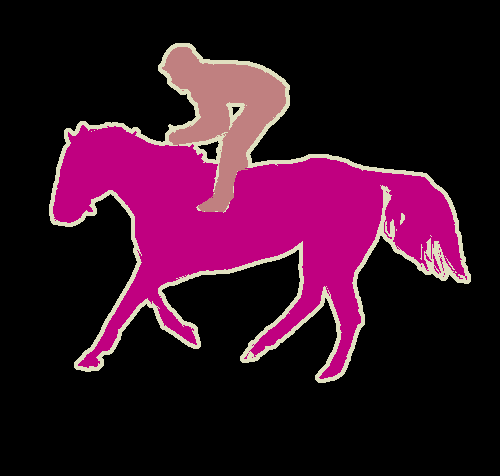} \\

\hline
Failure cases:\\
\includegraphics[width=.12\textwidth]{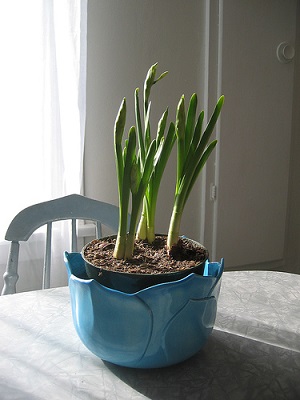} &
\includegraphics[width=.12\textwidth]{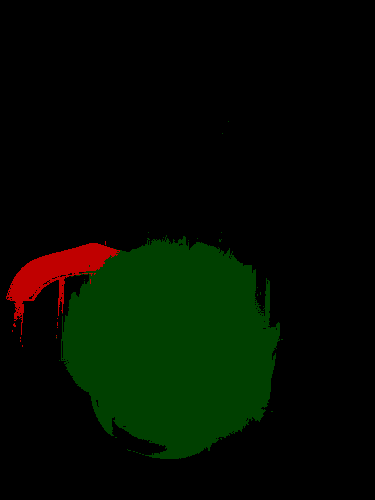} &
\includegraphics[width=.12\textwidth]{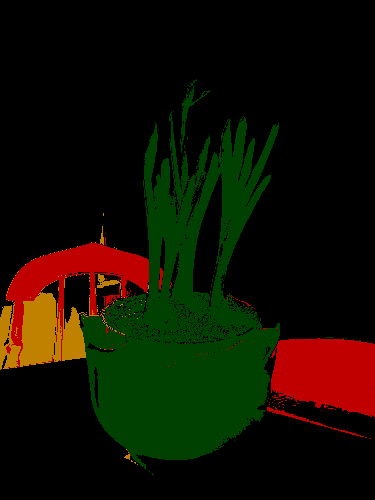} &
\includegraphics[width=.12\textwidth]{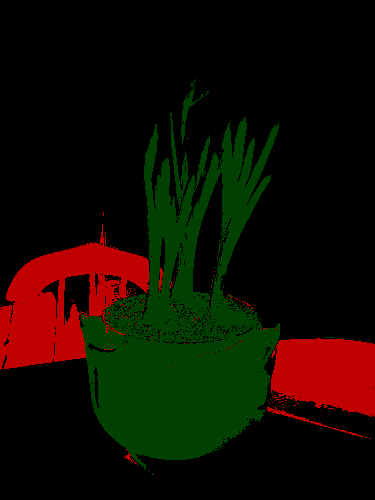} &
\includegraphics[width=.12\textwidth]{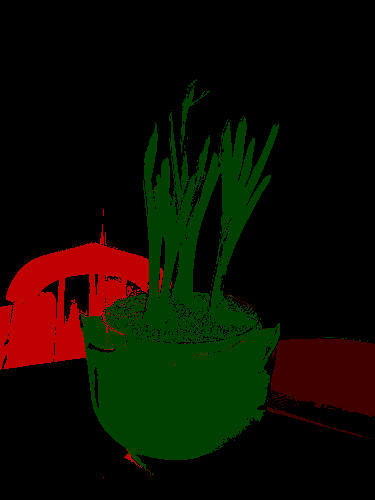} &
\includegraphics[width=.12\textwidth]{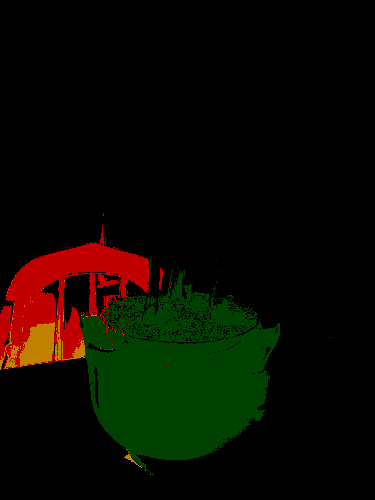} &
\includegraphics[width=.12\textwidth]{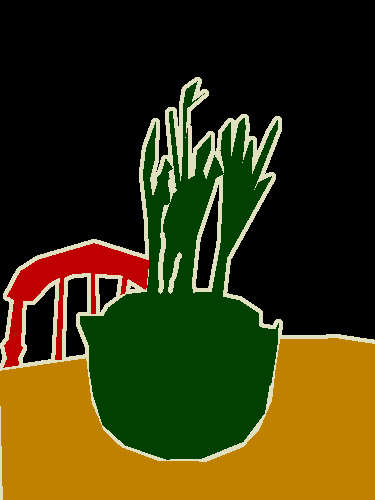} \\

\includegraphics[width=.12\textwidth]{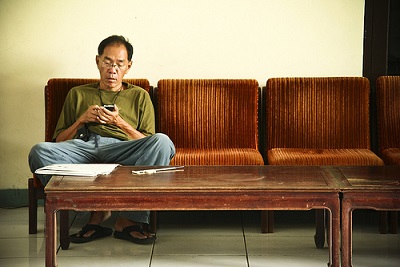} &
\includegraphics[width=.12\textwidth]{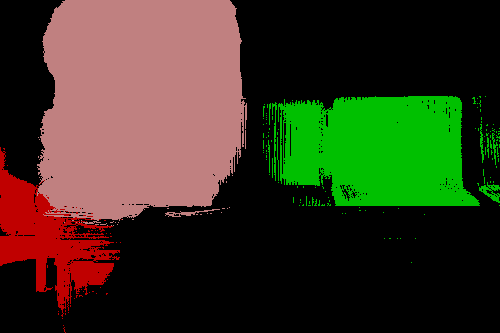} &
\includegraphics[width=.12\textwidth]{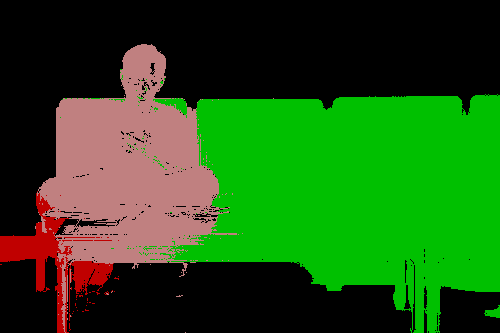} &
\includegraphics[width=.12\textwidth]{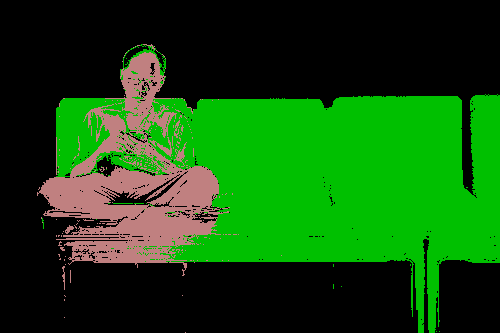} &
\includegraphics[width=.12\textwidth]{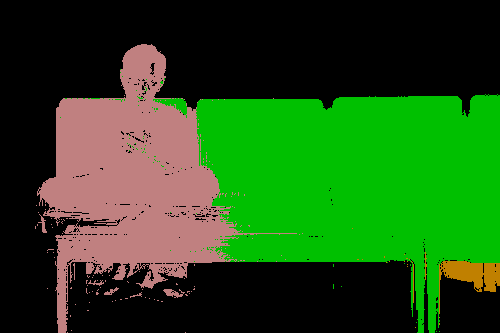} &
\includegraphics[width=.12\textwidth]{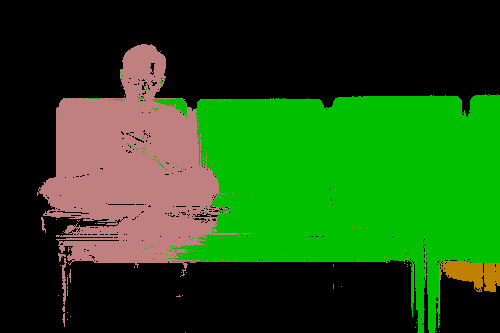} &
\includegraphics[width=.12\textwidth]{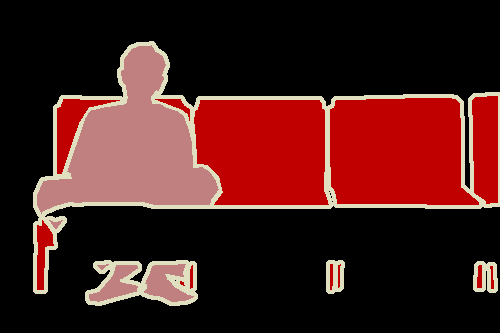} \\
\end{tabular}
\caption{Qualitative results: From left to right, there is image, the results of the model trained on Pascal VOC (column 2, 3, and 4), the results of the model trained on Flickr (column 5 and 6), and the groundtruth. The last two row shows the failure cases.}
\label{fig:results}
\end{figure}

We then further used our CheckMask procedure to evaluate how much can be gained by some cheap additional weak supervision. Note that, here, we were unable to report the result of the CCNN with additional supervision, since, in practice, we did not have access to per-image object size information.  Our results in Table~\ref{tab:flickr} evidence the benefits of our CheckMask procedure over our tag-only approach (see Fig.~\ref{fig:results} for qualitative results). Note that selecting the best mask for all 7238 training images took roughly 5 hours, which corresponds to 2.5 sec per image. This shows that our additional level of weak supervision remains very cheap to compute.
\begin{table}[t!]

\centering
\tiny
\caption{Per class IOU on the PASCAL VOC 2012 validation set for models trained with a subset of the MIRFLICKR-1M dataset.}
\label{tab:flickr}
\scalebox{0.84}
{
\begin{tabular}{| l|c c c c c c c c c c c c c c c c c c c c c|c |}
\hline
Method & \rotatebox[origin=c]{90}{bg} & \rotatebox[origin=c]{90}{aero} & \rotatebox[origin=c]{90}{bike} & \rotatebox[origin=c]{90}{bird} & \rotatebox[origin=c]{90}{boat} & \rotatebox[origin=c]{90}{bottle} & \rotatebox[origin=c]{90}{bus} & \rotatebox[origin=c]{90}{car} & \rotatebox[origin=c]{90}{cat} & \rotatebox[origin=c]{90}{chair} & \rotatebox[origin=c]{90}{cow} & \rotatebox[origin=c]{90}{table} & \rotatebox[origin=c]{90}{dog} & \rotatebox[origin=c]{90}{horse} & \rotatebox[origin=c]{90}{mbike} & \rotatebox[origin=c]{90}{person} & \rotatebox[origin=c]{90}{plant} & \rotatebox[origin=c]{90}{sheep}& \rotatebox[origin=c]{90}{sofa} & \rotatebox[origin=c]{90}{train} & \rotatebox[origin=c]{90}{tv} & \rotatebox[origin=c]{0}{mIOU} \\
 \hline
CCNN (tags)~\cite{CCNN} & 70.8 &  28.6 &  16.3 &  26.8 &  24.3 &   33.8 &  42.4 &  46.4 &  41.0 &    11.7 &  19.7 &   0 & 39.6  &31.0 &  44.2 &  42.3   & 24.3 & 34.0 &    0 &   37.8 &   25.4  & 30.5\\
CCNN (tags)+CRF~\cite{CCNN} & 75.4 &  32.8 & 13.1 &  32.1 & 27.3 &   32.9 &   41.3 &   45.6 &  43.3 & 10.7 &  16.5 &         0  &  44.8 &   37.3 &   47.2 &   46.2 &   27.1	& 36.8 &   0 &  38.8 &  26.9 & 32.2\\
\hline
Ours (baseline)& 69.8 & 28.2 & 17.2 & 25.9 & 20.6 & 29.2 & 39.2 & 45.4 & 39.2 & 10.9 & 16.9 & 15.0 & 39.2 & 30.3 & 40.8 & 36.9 & 25.8 & 28.6 & 21.2 & 34.2 & 28.6 & 30.6\\
\hline
Ours (Tags) & 75.2 & 52.8 & 15.3 & 46.4 & 33.1 & 44.3 & 51.4 & 46.1 & 47.9 & 15.1 & 23.1 & 27.0 & 53.8 & 34.5 & 47.1 & 40.7 & 20.3 & 43.3 & 18.2 & 45.6 & 19.3 & 38.1\\
Ours (Tags)+CRF  & 76.9 &   57.1 &  15.6 &  40.1 &  35.6 &   41.0 &  54.8  &  49.1 &    48.8 &   17.0 &   21.1 &   31.2 &   54.4 &    35.7 &   51.6 &   42.8 &    21.2 & 42.7 &  16.8 &   47.4 &   18.2 & 39.0
\\
Ours (CheckMask)  & 82.6  & 63.4  & 20.5  & 53.6  & 43.2  & 46.4   & 58.5  & 57.8    & 52.8  & 14.0  & 27.7   & 26.9   & 55.7   & 41.7  & 54.2  & 47.7  & 25.8	& 48.6  & 21.5   & 54.2  & 25.6 & 43.9\\
Ours (CheckMask)+CRF  & 83.8   & 68.2  & 17.2  & 57.7  & 46.4   & 48.7  & 62.1  & 59.4 & 56.2   & 14.7  & 29.3  & 28.8   & 59.3   & 45.6    & 57.5   & 50.6   & 27.4 & 53.4   & 23.2  &  56.5 & 26.4 & 46.3 \\

  \hline
\end{tabular}
}
\end{table}

\section{Conclusion}
We have introduced a Deep Learning approach to weakly-supervised semantic segmentation that leverages foreground/background masks directly extracted from our network pre-trained for the task of object recognition. Our experiments have shown that our approach outperforms the state-of-the-art methods when trained on image tags only. Furthermore, we have introduced a new level of weak supervision, consisting of selecting one mask among a set of candidates. This procedure can be achieve very easily, taking only roughly 2-3 seconds per image, and yields a further significant boost in accuracy. In the future, we intend to study if jointly training the foreground/background mask extraction procedure and the weakly-supervised segmentation network can further improve our results.

\clearpage

\bibliographystyle{splncs}
\bibliography{egbib}
\end{document}